\documentclass[final,12pt]{colt2021}
\title[Effective Dimension and Concentration]{Concentration of Non-Isotropic Random Tensors with Applications to Learning and Empirical Risk Minimization}
\usepackage{times}

\usepackage{caption}
\usepackage{indentfirst}
\usepackage{cleveref}

\newtheorem{defn}{Definition}
\newtheorem{thm}{Theorem}
\newtheorem{prop}{Proposition}

\newtheorem{cor}{Corollary}

\newtheorem{lem}{Lemma}
\newtheorem{hyp}{Assumption}

\newcommand{\Dom}{{\rm Dom}}
\newcommand{\dE}{\mathbb{E}}

\newcommand{\op}{{\rm op}}
\newcommand{\Tr}{{\rm Tr}}
\newcommand{\Lip}{{\rm Lip}}
\newcommand{\cB}{\mathcal{B}}

\newcommand{\cN}{\mathcal{N}}
\newcommand{\deff}{d_{ \rm eff}}

\newcommand{\cH}{\mathcal{H}}
\newcommand{\II}{1\!\!{\sf I}}
\newcommand{\dP}{\mathbb{P}}
\newcommand{\cS}{\mathcal{S}}
\newcommand{\cV}{\mathcal{V}}
\newcommand{\dR}{\mathbb{R}}
\newcommand{\R}{\mathbb{R}}
\newcommand{\eps}{\varepsilon}

\newcommand{\N}{\mathbb{N}}	
\newcommand{\NRM}[1]{{{\left\| #1\right\|}}} % ||1||% Operators
\newcommand{\esp}[1]{{{\mathbb{E}\left[ #1\right]}}} % ||1||% Operators

\DeclareMathOperator*{\argmin}{arg\,min}

\def\P{\mathbb P}
\def\E{\mathbb E}

\usepackage{dsfont}

\coltauthor{%
 \Name{Mathieu Even} \Email{mathieu.even@inria.fr} \\
 \addr INRIA, DI/ENS, PSL Research University, Paris, France
 \AND
 \Name{Laurent Massoulié} \Email{laurent.massoulie@inria.fr}\\
 \addr INRIA, DI/ENS, PSL Research University, MSR-Inria Joint Centre, Paris, France
}

\begin{document}

\maketitle
\begin{abstract}
Dimension is an inherent bottleneck to some modern learning tasks, where optimization methods suffer from the size of the data. In this paper, we study non-isotropic distributions of data and develop tools that aim at reducing these dimensional costs by a dependency on an effective dimension rather than the ambient one.
Based on non-asymptotic estimates of the metric entropy of ellipsoids -that prove to generalize to infinite dimensions- and on a chaining argument, our uniform concentration bounds involve an \emph{effective dimension} instead of the global dimension, improving over existing results.
We show the importance of taking advantage of non-isotropic properties in learning problems with the following applications: \emph{i)} we improve state-of-the-art results in statistical preconditioning for communication-efficient distributed optimization, \emph{ii)} we introduce a non-isotropic randomized smoothing for non-smooth optimization. Both applications cover a class of functions that encompasses empirical risk minization (ERM) for linear models.
\end{abstract}
%\mathieu{titre à trouver!}


%\section{Definitions and Notations}
%\input{1-definitions}
%\mathieu{intro: parler de dimension effective, de pbms où dimension apparait. Motiver notre forme de variable aléatoire à étudier. Motiver la sous-gaussianité. Dimension effective dans MNIST?Gaussian Width and Stable rank/dimension (Vershynin)}

\section{Introduction}

The sum of \emph{i.i.d.}~symmetric random tensors of order 2 and rank 1 (\emph{i.e.}~symmetric random matrices of rank 1) is studied in probability and statistics both for theoretical and practical interests, the most classical application being covariance estimation. The empirical mean of such matrices follows the Wishart distribution \citep{wishart1928,uhlig1994}.  \cite{Mar_enko_1967} proved the convergence in law of their spectrum when the number of observations and the dimension are of the same order. Machine Learning applications however require non-asymptotic properties, such as concentration bounds for a potentially large finite number of observations and finite dimension \citep{Tropp_2011,tropp2015introduction,donoho2017optimal,minsker2017extensions}, to control the eigenvalues of sums of independent matrices, namely:
\begin{equation}\label{eq:intro_norm_matrix}
    \NRM{\frac{1}{n}\sum_{i=1}^na_ia_i^\top - \E\left[aa^\top\right]}_\op = \sup_{\NRM{x}\le 1}\frac{1}{n}\sum_{i=1}^n x^\top \left(a_ia_i^\top -\E\left[aa^\top\right]\right)x
\end{equation} for $a,a_1,...,a_n$ \emph{i.i.d.}~random variables in $\R^d$. 

\subsection{Theoretical Contributions}

Our main contribution consists in new tools for the control of quantities generalizing \eqref{eq:intro_norm_matrix}. More precisely, for $r\ge2$, $f_1,...,f_r$ Lipschitz functions on $\R$, $a,a_1,...,a_n$ \emph{i.i.d.}~random variables in $\R^d$, and $\cB$ the $d$-dimensional unit ball, we derive in Section \ref{section:chaining} concentration bounds on:
\begin{equation}
    \sup_{x_1,\ldots,x_r\in \cB} \left \{ \frac{1}{n}\sum_{i\in[n]}\left(\prod_{k=1}^r f_k(a_i^\top x_k)- \E\left[\prod_{k=1}^r f_k(a^\top x_k)\right]\right)\right\}.\label{eq:intro_quantity}
\end{equation}
We thereby extend previous results in three directions.
\emph{i)} Matrices are tensors of order $2$, which we generalize by treating symmetric random tensors of rank 1 and order $r\ge 2$ (Section \ref{section:tensors}).
\emph{ii)} We consider non linear functions $f_i$ of scalar products $\langle a_i,x\rangle$, motivated by Empirical Risk Minimization. \eqref{eq:intro_quantity} can thus be seen as the uniform maximum deviation of a symmetric random tensor of order $r$ and rank 1, with non-linearities $f_1,...,f_r$.
\emph{iii)} Finally, by observing that data are usually distributed in a non-isotropic way (the \emph{MNIST} dataset lies in a 712 dimensional space, yet its empirical covariance matrix is of effective dimension less than 3 for instance), we generalize classical isotropic assumptions on random variables $a_i$ by introducing a non-isotropic counterpart:
\begin{defn}[$\Sigma$-Subgaussian Random Vector] A random variable $a$ with values in $\R^d$ is $\Sigma$ -subgaussian for $\Sigma\in \R^{d\times d}$ a positive-definite matrix if:
\begin{equation}
    \forall t>0,\forall x\in \cB, \P(|a^\top x|>t)\le 2\exp\left(-\frac{1}{2}\frac{t^2}{ x^\top \Sigma x}\right).\label{eq:subgauss}
\end{equation}\end{defn}
A gaussian $\cN(0,\Sigma)$ is for instance $\Sigma$-subgaussian. Note however that in the general case, $\Sigma$ is not equal to the covariance matrix. The aim is then to derive concentration bounds on \eqref{eq:intro_quantity} (Section \ref{section:chaining}) that involve an \emph{effective dimension} of $\Sigma$: a quantity smaller than the global dimension $d$, that reflects the non-isotropic repartition of the data:
\begin{defn}[Effective Dimension $\deff(r)$] Let $\Sigma\in\R^{d\times d}$ a symmetric positive semi-definite matrix of size $d\times d$, where $d\in \N^*$. Let $\sigma_1^2\ge\sigma_2^2\ge...\ge\sigma_d^2\ge0$ denote its ordered eigenvalues. For any $r\in \N^*$, let $\deff(r)$ be defined as follows:
\begin{equation}
    \label{eq:deff_r}
    \deff(r):=\sum_{i=1}^d\left(\frac{\sigma_i}{\sigma_1}\right)^{\frac{2}{r}}=\frac{\Tr (\Sigma^{1/r})}{\NRM{\Sigma^{1/r}}_\op}.
\end{equation}
This notion generalizes \emph{intrinsic dimension} in \cite{tropp2015introduction} and \emph{stable rank} in \cite{vershynin2011introduction,vershynin2014estimation}, both obtained for $r=1$.
\end{defn}

\noindent \textbf{Chaining Argument and Metric Entropy of Ellipsoids:} Control of \eqref{eq:intro_quantity} involves a chaining argument (\cite{boucheron:hal-00942704}, Chapter 13). In the simplest version of chaining, in order to bound a random variable of the form $\sup_{t\in \mathcal{T}}X_t$, one discretizes the set of indices $\mathcal{T}$ and approximates the value $\sup_{t\in \mathcal{T}}X_t$ by a supremum taken over successively refined discretizations.
%In \cite{hendrikx2020statistically}, a weaker version of Theorem \ref{thm:chaining_centered} involving the global dimension $d$ instead of $\deff(r)$ is proved, by covering the unit ball $\cB$ with a sequence of nested balls of radii $(\eps_j)_{j\ge0}$ tending to zero. The dependency in the global dimension comes from the fact that the size of an $\eps$-net of balls of radius $\eps$ covering the unit ball, is of order $(1+2\eps^{-1})^d$.
To exploit the non-isotropic properties of $\Sigma$-subgaussian random variables, we apply chaining based on a covering of the unit ball $\cB$ with ellipsoids. Our approach yields similarities with that of \citet{zhong2017eigenvector}, who uses chaining with ellipsoids for a different purpose (control of eigenvectors).
In section \ref{section:tensors}, in the setting where $f_1=...=f_r=Id$, control of \eqref{eq:intro_quantity} reduces to controlling the operator norm of empirical tensors. This can be done using our bounds on the $\eps$-entropy of ellipsoids, without the use of chaining.

In Section \ref{section:ellips}, we present results  on the number of balls of fixed radius $\eps$ needed to cover an ellipsoid in dimension $d$. The logarithm of this quantity is often called the $\eps$\emph{-entropy} of an ellipsoid. \cite{Dumer2004OnCO} studied the limit $d\to\infty$, while we provide non-asymptotic estimates.  Furthermore, in Appendix \ref{app:ellips_infinite}, we extend these results to ellipsoids in infinite dimension, obtaining bounds on metric entropy in terms of power-law norm decay.We believe these technical results (both in finite and infinite dimension) to be of strong practical and theoretical interests: the bridge between covering numbers and suprema of random subgaussian processes is rather thin due to Dudley's inequality \citep{DUDLEY1967290}. Bounding metric entropy of ellipsoids is thus a step towards uniform bounds on more general random variables than the one we consider in \eqref{eq:intro_quantity}.

\subsection{Applications in Learning Problems and ERM}
We show the relevance of our concentration bounds through the following applications.

\paragraph{Operator Norm Of Tensors} Setting $f_1=...=f_r=Id$ yields the operator norm of the empirical tensor $\frac{1}{n}\sum_i a_i^{\otimes r} - \E a^{\otimes r}$ in \eqref{eq:intro_quantity}. In Section \ref{section:tensors} we derive precise large deviation bounds on such tensors involving the effective dimension $\deff(r)$, improving on previous works \citep{bubeck2020law,paouris2017dvoretzky} which depended on the global dimension. Optimal concentration inequalities on rank 1 symmetric tensors (\emph{i.e.}~of the form $a^{\otimes r}$) are not known. We refer the interested reader to \citet{vershynin2020tensors} for the study of rank 1 tensors of the form $a_1\otimes...\otimes a_r$ where $a_1,...,a_r$ are \emph{i.i.d.}~random variables, a different problem than ours.
In Appendix \ref{F-robustness}, we apply these bounds to the study of the Lipschitz constant of two-layered neural networks with polynomial activation, elaborating on the results in \cite{bubeck2020law}.

\paragraph{Concentration of Hessians and Statistical Preconditioning} For $\ell$ a twice differentiable function on $\R$ and Hessian-Lipschitz, let $f(x)=\frac{1}{n}\sum_{i=1}^n\ell(a_i^\top x)$. Then, $\nabla^2f(x)=\frac{1}{n}\sum_{i=1}^n \ell''(a_i^\top x)a_ia_i^\top$, and setting $r=3,f_1=\ell'',f_2=f_3=Id$ in \eqref{eq:intro_quantity} yields $\sup_{x\in \cB}\NRM{\nabla^2 f(x) - \E [\nabla^2f(x)]}_\op$. Controlling such quantities is relevant in optimization when studying functions that have an empirical risk structure. Methods such as statistical preconditioning \citep{shamir2014communication} take advantage of the \emph{i.i.d.}~structure of the observations, as we illustrate in Section \ref{section:precond}. Our results improve on the state of the state-of-the-art \citep{hendrikx2020statistically}, establishing guarantees based on $\deff(r)$ rather than~$d$.

\paragraph{Randomized Smoothing} Minimizing a non-smooth convex function $f$ is a difficult problem, as acceleration methods cannot be used. \cite{duchi2012randomized,scaman2018optimal} propose to use the gradients of $f^\gamma$ a smoothed version of $f$, where $f^\gamma(x)=\E_{X\sim\cN(0,I_d)}[f(x+\gamma X)]$. This method suffers from a dimensional cost, a factor $d^{1/4}$ in the convergence time, that cannot in general be removed \citep{bubeck2019complexity,nemirovsky1985}. In Section \ref{section:smooth}, considering an empirical risk structure for $f$ and a non-isotropic smoothing distribution for $X$, we take advantage of the non-isotropic repartition of data to obtain an effective dimension $\deff(r)$ instead of the whole dimension~$d$.

\paragraph{Organization of the paper} We first present our 3 main uniform concentration bounds in Section \ref{section:chaining}: control of \eqref{eq:intro_quantity} and of the same quantity but un-centered (both using chaining), and a more precise control of \eqref{eq:intro_quantity} in the case where $f_1=...=f_r=Id$ (control of empirical mean of symmetric random tensors of rank 1 ann order $p$). In section \ref{section:ellips}, we provide bounds on the metric entropy of ellipsoids in terms of effective dimension. We also investigate the case of infinite dimension with the notion of spectral dimension. The last two sections present two applications of the results presented in Section \ref{section:chaining}. In Section \ref{section:precond}, we apply Theorem \ref{thm:chaining_centered} to control uniform deviation of Hessians, in order to prove that statistical preconditioning methods naturally adapt to the underlying effective dimension. In Section \ref{section:smooth}, we introduce a non-isotropic smoothing method for empirical risk minimization.

\section{Main Theoretical Results\label{section:chaining}}
\subsection{Concentration Bound With Centering}

\begin{thm}[Concentration With Centering] \label{thm:chaining_centered}Let $r\ge2$ and $d,n\ge1$ integers. Let $\Sigma\in \R^{d\times d}$ a positive-definite matrix and $a,a_1,...,a_n$ \emph{i.i.d.}~$\Sigma-$subgaussian random variables. Let $\deff(s),s\in\N^*$ be defined as in \eqref{eq:deff_r}. Let $f_1,...,f_r$ be 1-Lipshitz continuous functions on $\R$ such that $f_i(0)=0$ for $i\in [n]$. For all $k=1,...,r$, let $B_k>0$ such that:
\begin{equation}\label{eq:hyp_thm1}
    \forall x\in \cB,\forall i\in [n], |f_k(a_i^\top x)|\leq B_k \text{ almost surely.}
\end{equation}
Let $B=B_1...B_k$. Define the following random variable:
\begin{equation}
    \label{eq:var_with_centering}
Z:=\sup_{x_1,\ldots,x_r\in \cB} \left \{ \frac{1}{n}\sum_{i\in[n]}\left(\prod_{k=1}^r f_k(a_i^\top x_k)- \E\left[\prod_{k=1}^r f_k(a^\top x_k)\right]\right)\right\}.
\end{equation}
Then, for any $\lambda>0$ and for some universal constant $C_r$, the following large-deviation bound holds:
\begin{equation}\label{eq:high_dev_without_centering}
    \dP\left(Z\ge C_r\sigma_1^r \left(\frac{1}{n}\frac{\lambda+ \deff(r)\ln(d)}{\left(\sigma_1^{-r}B\right)^{2/r-1}} +  \frac{\sqrt{\lambda}+\sqrt{\deff(1)\ln(d)}}{\sqrt{n}}\right)\right)\le e^{-\lambda}.
\end{equation}
\end{thm}

\subsection{Concentration Bound Without Centering}

\begin{thm}[Concentration Without Centering] \label{thm:chaining_not_centered}Let $r\ge2$ and $d,n\ge1$ integers. Let $\Sigma\in \R^{d\times d}$ a positive-definite matrix and $a,a_1,...,a_n$ \emph{i.i.d.}~$\Sigma-$subgaussian random variables \eqref{eq:subgauss}. Let $\deff(s),s\in\N^*$ be defined as in \eqref{eq:deff_r}. Let $f_1,...,f_r$ be 1-Lipshitz continuous functions on $\R$ such that $f_i(0)=0$ for $i\in [n]$. For all $k=1,...,r$, let $B_k>0$ such that:
\begin{equation*}\label{eq:hyp_thm2}
    \forall x\in \cB,\forall i\in [n], |f_k(a_i^\top x)|\leq B_k \text{ almost surely.}
\end{equation*}
Let $B=B_1...B_k$. Define the following random variable:
\begin{equation}
    \label{eq:var_without_centering}
Y:=\sup_{x_1,\ldots,x_r\in \cB} \frac{1}{n}\sum_{i\in[n]}\prod_{k=1}^r f_k(a_i^\top x_k).
\end{equation}
Then, for any $\lambda>0$ and for some universal constant $C_r$, the following large-deviation bound holds:
\begin{equation}\label{eq:high_dev_without_centering}
    \dP\left(Y\ge \sigma_1^r C_r\left(1+\frac{\deff(r)\ln(d)+\lambda}{n}(\sigma_1^{-r} B)^{1-2/r}\right)\right)\le e^{-\lambda}.
\end{equation}
\end{thm}
\begin{remark}\label{rq:non_centered_thm} Assumptions \eqref{eq:hyp_thm1} in Theorem \ref{thm:chaining_centered} can be replaced by high-probability bounds on the random variables $a_1,...,a_n$ in the following way.
If we denote $R=\sup_{i=1,...,n}\NRM{a_i}$, we always have $B_k\le R$ and $B\le R^r$ using Lipshitz continuity of functions $f_k$. Furthermore, a Chernoff bound gives with probability $1-\delta$: 
\begin{equation*}R^2\le 4\sigma_1^2(2\deff(1)+\ln(1/\delta)+\ln(n)),
\end{equation*}
yielding, with probability $1-\delta$, where $Z$ is defined in \eqref{eq:var_with_centering}:
\begin{equation*}
    Z\leq C_r\sigma_1^r\left(\frac{\ln(\delta^{-1})+\deff(r)\ln(d)}{n}\big(\deff(1)+\ln(\delta^{-1})+\ln(n)\big)^{\frac{r}{2}-1} + \frac{\sqrt{\ln(\delta^{-1})}+\sqrt{\deff(1)\ln(d)}}{\sqrt{n}}\right).
\end{equation*}
The same reasoning applies to Theorem \ref{thm:chaining_not_centered} in the case without centering.
\end{remark}
\begin{remark}
Theorems \ref{thm:chaining_centered} and \ref{thm:chaining_not_centered} assume that the functions $f_k$ are 1-Lipshitz and that the supremum is taken over $\cB$ the centered unit ball. By considering $L_k$-Lipshitz functions and a ball $\cB(x_0,\rho)$, one obtains the same bound, up to a factor $\rho L_1...L_r $.
\end{remark}
\begin{remark}
In Appendix \ref{app:tightness}, we study the tightness of these results.  We prove that, for $f_1=...=f_r=I_d$:
\begin{equation*}
  \left\{  \begin{array}{ccc}
        \E[Y_{{\rm Non-Centered }}] &\ge& C_1\sigma_1^r\left( 1+\frac{\deff(1)^{r/2}}{n}\right)  \\
        \E[Y_{{\rm Centered }}^2] & \ge & C_2\sigma_1^{2r}\frac{C(\Sigma)}{n}.
    \end{array}\right.
\end{equation*}Dependency in terms of $n$ is thus optimal in both Theorems  \ref{thm:chaining_centered} ($O(1/\sqrt{n})$) and  \ref{thm:chaining_not_centered} ($\sigma_1^r(1+O(1/n))$). 
However, we believe that both the factor $\ln(d)$ and having $\deff(r)$ instead of $\deff(1)$ are artifacts of the proof, coming from our non-asymptotic estimates of the metric entropy of ellipsoids (next Section).
\end{remark}

\subsection{Concentration of Non-Isotropic Random Tensors\label{section:tensors}}
In this section, we provide a concentration bound on the empirical mean of symmetric random tensors of rank 1, involving an effective dimension. In the appendix, we exploit this result to derive some results on the robustness of two-layered neural networks with polynomial activations (Appendix \ref{F-robustness}). Methods such as in \cite{paouris2017dvoretzky,bubeck2020law}, which do not rely ellipsoids, cannot yield results as sharp as ours, as detailed in Appendix \ref{C-Tensors}.

%and our chaining method. However, bounds involving the full dimension do not require any chaining method, which is not the case here. We hence emphasize the necessity of chaining in Appendix \ref{C-Tensors}, where we provide two other large deviation propositions without chaining: one involving the global dimension, another requiring a condition between $p$ and the effective dimension.
\begin{defn}[Tensor] A tensor of order $p\in\N^*$ is an array $T=(T_{i_1,...,i_p})_{i_1,...,i_p\in [d]}\in \R^{dp}$.

\noindent $T$ is said to be of rank $1$ if it can be written as:
$$T=u_1\otimes \cdots \otimes u_p$$ for some $u_1,...,u_p \in \R^p$.

\noindent Scalar product between two tensors of same order $p$ is defined as:
\begin{equation*}
    \langle T, S \rangle = \sum_{i_1,...,i_p} T_{i_1,...,i_p}S_{i_1,...,i_p}\text{, giving the norm: } \NRM{T}^2=\sum_{i_1,...,i_p}T_{i_1,...,i_p}^2.
\end{equation*}
\noindent We define the operator norm of a tensor as:
\begin{equation*}
    \NRM{T}_{\op}=\sup_{\NRM{x_1\otimes...\otimes x_p}\le 1} \langle T, x_1\otimes...\otimes x_p \rangle.
\end{equation*}
\end{defn}

\begin{defn}[Symmetric Random Tensor of Rank 1]
A symmetric random tensor of rank 1 and order $p$ is a random tensor of the form:
\begin{equation}
    T=X^{\otimes p},
\end{equation}
where $X\in \R^d$ is a random variable. We say that $T$ is $\Sigma$-subgaussian if $X$ is a $\Sigma$-subgaussian random variable.
\end{defn}

\noindent We wish to bound the operator norm of tensors of the form $T=\frac{1}{n}\sum_{i=1}^nT_i$, where $T_1,...,T_n$ are \emph{i.i.d.}~subgaussian random tensors of rank 1 and order $p$, using a dependency in an effective dimension rather than the global one. 
We have:
\begin{align*}
    \NRM{T-\E T}_\op    & =  \frac{1}{n} \sup_{x_1,...,x_p \in \cS} \sum_{i=1}^n \left\{\prod_{k=1}^p\langle a_i,x_k\rangle - \esp{\prod_{k=1}^p\langle a_i,x_k\rangle}\right\}.
\end{align*}
This quantity can be upper-bounded using chaining as in Theorem \ref{thm:chaining_centered}. However, using a simpler argument (Appendix \ref{app:tensors}) inspired by \citet{bubeck2020law} and our bounds on the metric entropy of ellipsoids, we have the following.
\begin{thm}[Non-Isotropic Concentration Bound on Random Tensors]\label{prop:tensors_non_iso_main}
Let $T_1,...,T_n$ be \emph{i.i.d.} random tensors of order $p$, rank $1$, symmetric and $\Sigma$-subgaussian. Let $T=\frac{1}{n}\sum_{i=1}^n T_i$. With probability $1-\delta$ for any $\delta>0$ and universal constant $C_p>0$, we have:
\begin{align*}
    \NRM{T-\E T}_{\op} \leq C_{p}\sigma_1^p\sqrt{\frac{\big(\deff+\ln(d)+\ln(\delta^{-1})\big)^{p+1}\ln(n)^p}{n}}.
\end{align*}
Equivalently, for any $\lambda>0$:
\begin{equation*}
    \P\left(\NRM{T-\E T}_\op \geq C_{p}\sigma_1^p\sqrt{\frac{\big(\deff+\ln(d)+\lambda\big)^{p+1}\ln(n)^p}{n}}\right)\leq e^{-\lambda}.
\end{equation*}
\end{thm}

\noindent This concentration bound is similar to those of \cite{zhivotovskiy2021dimensionfree,vershynin2011tensors,GUEDON2007798}, up to logarithmic factors we lose in our approach. Yet, even though our proof is quite straightforward, we obtain results similar to some using much more involved and sophisticated arguments.
We next present the upper-bounds on the number of balls needed to cover an ellipsoid we use to prove the concentration bounds with chaining (Theorems \ref{thm:chaining_centered} and \ref{thm:chaining_not_centered}) or without (Theorem \ref{prop:tensors_non_iso_main}). We believe these technical results to be of independent interest due to the strong link between metric entropy and uniform concentration bounds.

\section{Results on Covering of Balls with Ellipsoids and Metric Entropy\label{section:ellips}}

\subsection{Metric Entropy of an Ellipsoid}

\begin{defn}[Ellipsoid and $\eps$-Entropy] \label{def:ellips1}
Given a vector $b=(b_1,\ldots,b_d)$ with $b_1\ge \cdots \ge b_d>0$, the ellipsoid $E_b$ is defined as
$$
E_b=\left\{x\in\dR^d: \sum_{i\in[d]}\frac{x_i^2}{b_i^2}\le 1\right\}.
$$
The $\eps$-\emph{entropy} $\cH_\eps(E_b)$ of ellipsoid $E_b$ is the logarithm of the size of a minimal $\eps$-covering (or $\eps$-net in information theory terminology) of $E_b$. More formally:
\begin{equation}
    \cH_\eps(E_b)=\ln\left( \min\left\{|A|: A\subset \R^d, E_b\subset \bigcup_{x\in A} \cB(x,\eps)\right\}\right),
\end{equation}
where $\cB(x,\eps)$ is the Euclidean ball of radius $\eps$. The unit entropy is the $\eps$-entropy for $\eps=1$.
\end{defn}
Given an ellipsoid $E_b$, define the following quantities:
\begin{equation}
    K_b=\sum_{i=1}^{m_b}\ln(b_i) \text{ and } m_b=\sum_{i\in[d]}\II_{b_i>1}.
\end{equation}
Provided that:
\begin{equation}
    \ln(b_1)=o\left(\frac{K_b^2}{m_b\ln(d)}\right),\label{eq:dumer_result}
\end{equation}\cite{Dumer2004OnCO} (Theorem 2 in their article) prove the following asymptotic equivalent of $\cH_1(E_b)$ when $d\to \infty$:
\begin{equation}
    \cH_1(E_b)\sim K_b.
\end{equation}
However, we need non-asymptotic bounds on $\cH_1(E_b)$. Using techniques introduced in \cite{Dumer2004OnCO}, we thus establish Theorem \ref{thm:ellipsoid_non_asympt}, whose proof appears in Appendix \ref{section:appedix_ellipsoids}, together with an extension to ellipsoids in infinite dimension.
%However, we are only interested in estimates for fixed dimensions, hence our analysis in Appendix \ref{section:appedix_ellipsoids}: our contributions in this area are clean and non-asymptotic bounds on $\cH_1(E_b)$ (Theorem \ref{thm:ellipsoid_non_asympt}) based on the proof technique of \cite{Dumer2004OnCO} and technical lemmas (Corollary \ref{cor1}). In Appendix \ref{app:ellips_infinite}, we study such ellipsoids in infinite dimension, and their link with spectral dimensions.
\begin{thm}[Unit Entropy of an Ellipsoid in Fixed Dimension]\label{thm:ellipsoid_non_asympt}
One has, for some universal constant $c>0$, the following bound on the unit entropy of ellipsoid $E_b$:
$$
\cH_1(E_b)\le K_b +c\left[\ln(d)+\sqrt{\ln(b_1) m_b\ln(d)}\right].
$$
\end{thm}
\noindent This theorem gives the following corollary, bounding the number of ellipsoids required to cover the unit ball, directly linked with the number of balls required to cover an ellipsoid thanks to a linear transformation.

\subsection{Coverings of the Unit Ball With Ellipsoids}

\begin{cor}\label{cor1}Let $\eps>0$.
Let random vector $a\in\dR^d$ satisfy subgaussian tail assumption \eqref{eq:subgauss} for matrix $\Sigma$, with spectrum $\sigma_1^2\ge \cdots\ge \sigma_d^2>0$.
Then there exists a collection $\cN_\eps$ of vectors in $\cS_1$ the unit sphere of $\R^d$ such that, for all $x\in \cS_1$, there exists $y=\Pi_\eps x\in\cN_\eps$ such that 
\begin{equation}\label{eq:cor1}
\|x-y\|_{\Sigma}^2:=(x-y)^\top \Sigma (x-y)\le \eps^2\sigma_1^2,
\end{equation}
and the covering $\cN_\eps$ verifies
\begin{equation}\label{eq:cor2}
\ln(|\cN_\eps|)\le \cH_\eps:=\sum_{i=1}^{m_\eps}\ln\left(\frac{\sigma_i}{\eps\sigma_1}\right) +c\left[\ln(d)+\sqrt{\ln(\eps^{-1})\ln(d)m_\eps}\right],
\end{equation}
where 
\begin{equation}\label{eq:cor3}
m_\eps=\sum_{i=1}^d\II_{\sigma_i>\eps\sigma_1}
\end{equation}
and $c$ is some universal constant. Furthermore, we have:
\begin{equation*}
    \cH_\eps\leq \ln(\eps^{-1})+\frac{\min\left(d-1,\eps^{-\frac{2}{r}}\frac{\deff(r)-1}{e}\right)}{2/r}\ln\left(\max(e,\eps^{-\frac{2}{r}}\frac{\deff(r)-1}{d-1})\right)+c\left [\ln(d) +\sqrt{\ln(\eps^{-1})\ln(d)m_\eps}\right],
\end{equation*}
and 
\begin{equation*}
    m_\eps\leq 1+(\deff(r)-1) \eps^{-\frac{2}{r}}.
\end{equation*}
\end{cor}
This last bound on $\cH_\eps$ is a core technical lemma behind Theorems \ref{thm:chaining_centered} and \ref{thm:chaining_not_centered} . It is to be noted that $\cH_\eps$ is not linear in an effective dimension. Indeed, for $\eps\le C_r \left(\frac{d-1}{\deff(r)-1}\right)^{r/2}$, our expression is linear in $d$. This difficulty is the non-asymptotic equivalent of \cite{Dumer2004OnCO}'s assumption in~\eqref{eq:dumer_result}.

\subsection{Ellipsoids in Infinite Dimension}

We here define ellipsoids in infinite dimension and upper-bound asymptotically their $\eps$-entropy in terms of \emph{spectral dimension}. Although not used in the applications described in the present article, uniform concentration of infinite-dimensional random vectors that satisfy an infinite-dimensional subgaussian property require results such as the one we provide below.

Let $\cV$ be a separable real Hilbert space (\emph{e.g.}~$\R^{\N}$, $\ell^2([0,1])$). 
\begin{defn}[Ellipsoids in Hilbert Spaces] 
Let $A$ a self-adjoint and semi-definite positive operator on $\cV$ \emph{i.e.}~such that $\forall (x,y)\in\cV^2$, we have $\langle A(x),y\rangle=\langle x,A(y)\rangle\geq0$. We define the ellipsoid $E_A\subset \cV$ by:
\begin{equation*}
    E_A=\left\{x\in\cV :\NRM{A^{\dagger}(x)}^2\leq 1\right\},
\end{equation*}
where $A^{\dagger}$ is the pseudo-inverse of $A$.
\end{defn}
This notion generalizes Definition \ref{def:ellips1}: taking $\cV=\R^d$ and $A={\rm Diag}(b_1,...,b_d)$, we have $E_A=E_b$. We next define the spectral dimension of an ellipsoid. We recall that if $A$ is a self-adjoint and semi-definite positive operator on $\cV$, there exists a Hilbert basis of eigenvectors of $A$, and the eigenvalues of $A$ are non-negative.
\begin{defn}[Spectral Dimension and Effective Dimension] Let $E_A$ an ellipsoid in $\cV$, where $A$ is a self-adjoint and semi-definite positive operator. Assume that the eigenvalues of $A$ can be ordered as a decreasing sequence~$(b_i)_{i\in\N^*}$. $E_A$ is of spectral dimension $d\in\R^{+,*}$ if $\sum_{i\in\N^*}b_i^2 <\infty$ and when~$n\to\infty$:
\begin{equation*}
    \sum_{i\geq n+1} b_i^2 = \mathcal{O}\big(n^{-\frac{2}{d}}\big).
\end{equation*}
The effective dimension of ellipsoid $E_A$ is then $\sum_{i\in\N^*}b_i^2$.
\end{defn}
The right notion of dimension for the control of metric entropy in infinite dimension is the spectral dimension, as shown in the next proposition: the $\eps$-entropy of an ellipsoid scales as the spectral dimension in infinite dimension.
\begin{prop}
Let $E_A$ be an ellipsoid in $\cV$, of spectral dimension $d>0$. We have, when $\eps\to0$:
\begin{equation*}
    \cH_{\eps}(E_A)\leq d\ln\big(\eps^{-1}\big)^2\big(1+o(1)\big),
\end{equation*}
where $\cH_{\eps}(E_A)$ is the number (possibly infinite) of balls of radius $\eps$ required to cover $E_A$.
\end{prop}

\section{Statistical Preconditioning: Bounding Relative Condition Numbers\label{section:precond}}
In this section, we present an application of Theorem \ref{thm:chaining_centered} to optimization. Essentially, we show that statistical preconditioning-based optimization automatically benefits from low effective dimension in the data, thus proving a conjecture made in \citet{hendrikx2020statistically}.

%, and more precisely to statistical preconditioning. The result we obtain at the end of this section was conjectured by \cite{hendrikx2020statistically}. We show with strong theoretical guarantees that one can take advantage of a small effective dimension over a much larger global one while optimizing.

\subsection{Large Deviation of Hessians}

Let $f$ be a convex function defined on $\R^d$. We assume that the following holds, which is true for logistic or ridge regressions (Appendix \ref{app:ERM}).
\begin{hyp}[Empirical Risk Structure]\label{hyp:erm}Let $\ell:\R\to \R$ convex, twice differentiable such that $\ell''$ is $\NRM{\ell''}_{\Lip}$-Lipschitz.
Let $n\in\N^*$, some convex functions $\ell_{j}:\R\to\R,j\in [n]$ such that $\forall j\in[n], \ell_{j}''=\ell''$ and \emph{i.i.d.}~$\Sigma$-subgaussian random variables $(a_{j})_{j\in[n]}$. We assume that:
\begin{equation}
    \forall x\in \R^d, f(x)=\frac{1}{n}\sum_{j=1}^{n}\ell_{j}(a_{j}^\top x).
\end{equation}
\end{hyp}
\begin{prop} \label{prop:deviation_hess}Denote $H_x$ the Hessian of $f$ at some point $x\in\R^d$ and $\Bar{H}_x$ its mean. We have:
\begin{equation*}
    H_x=\frac{1}{n}\sum_{i=1}^n \ell''(a_i^\top x)a_ia_i^\top, \text{ and } \Bar{H}_x=\E_a\left[\ell''(a_1^\top x)a_1a_1^\top\right].
\end{equation*}
Let:
\begin{equation}
    Z=\sup_{\NRM{x}\le 1} \NRM{H_x-\Bar{H}_x}_{\op}.
\end{equation}With probability $1-\delta$, we have, with $C$ a universal constant:
\begin{equation*}
    Z \leq C \sigma_1^3\NRM{\ell''}_{\Lip}\left(\frac{(\deff(3)\ln(d)+\ln(1/\delta))\sqrt{\deff(1)+\ln(n/\delta)}}{n}+\frac{\sqrt{\ln(1/\delta)}+\sqrt{\deff(1)\ln(d)}}{\sqrt{n}}\right).
\end{equation*}
\end{prop}

\noindent Previous works \citep{hendrikx2020statistically} obtained:
\begin{equation}
    C' \sigma_1^3\NRM{\ell''}_{\Lip}\frac{(d+\ln(1/\delta))\sqrt{\deff(1)+\ln(n/\delta)}}{\sqrt{n}}\left[ \frac{1}{\sqrt{d}}+\frac{1}{\sqrt{n}}\right].
\end{equation}
In order for this bound to be of order $1$, $n$ was required to be of order the whole dimension $d$, while we only need $n$ to be of order $\deff(3)$. 

\subsection{Statistical Preconditioning}

Consider the following optimization problem:
\begin{equation}
    \min_{x\in\R^d} \Phi(x):= F(x)+\psi(x), 
\end{equation}
where $F(x)=\frac{1}{n}\sum_{j=1}^n f_j(x)$ has a finite sum structure and $\psi$ is a convex regularization function. 
%We further assume that we have $m$ workers, an each function $f_j$ is stored at worker $j$. A central unit may coordinate the workers/machines. 
Standard assumptions are the following:\begin{equation}\forall x, \sigma_FI_d\le \nabla^2F(x)\le L_F I_d .\end{equation} 
We focus on a basic setting of distributed optimization. At each iteration $t=0,1,...$, the server broadcasts the parameter $x_t$ to all workers $j\in\{1,...,n\}$. Each machine $j$ then computes in parallel $\nabla f_j(x_t)$ and sends it back to the server, who finally aggregates the gradients to form $\nabla F(x_t)=\frac{1}{n} \sum_j \nabla f_j(x_t)$ and use it to update $x_t$ in the following way, using a standard proximal gradient descent, for some parameter $\eta_t\le 1/L_F$:
\begin{equation}
    x_{t+1}\in \argmin_{x\in \R^d} \left \{ \langle \nabla F(x_t),x \rangle + \psi(x) +\frac{1}{2\eta_t} \NRM{x-x_t}^2 \right \}.\label{eq:proximal_gradient_descent}
\end{equation}
Setting $\eta_t=1/L_F$ yields linear convergence:
\begin{equation}
    \Phi(x_t)-\Phi(x^*)\le L_F (1-\kappa_F^{-1})^t\NRM{x_0-x^*}^2.
\end{equation}In general, using an accelerated version of \eqref{eq:proximal_gradient_descent}, one obtains a communication complexity (\emph{i.e.}~number of steps required to reach a precision $\eps>0$) of $O(\kappa_F^{1/2}\ln(1/\eps))$ (where $\kappa_F=\frac{L_F}{\sigma_F}$) that cannot be improved in general.
Statistical preconditioning is then a technique to improve each iteration's efficiency, based on the following insight: considering \emph{i.i.d.}~datasets leads to statistically similar local gradients $\nabla f_j$. The essential tool for preconditioning is the Bregman divergence.
\begin{defn}[Bregman divergence and Relative Smoothness]
For a convex function $\phi:\R^d\to\R$, we define $D_\phi$ its Bregman divergence by:
\begin{equation}
    \forall x,y\in \R^d,D_\phi(x,y)=\phi(x)-\phi(y)-\langle \nabla \phi(y),x-y\rangle.
\end{equation}
For convex functions $\phi,F:\R^d\to\R$, we say that $F$ is relatively $L_{F/\phi}$-smooth and $\sigma_{F/\phi}$-strongly-convex if, for all $x,y\in\R^d$:
\begin{equation}
    \sigma_{F/\phi}D_\phi(x,y)\le D_F(x,y)\le L_{F/\phi}D_\phi(x,y),
\end{equation}
or equivalently:
\begin{equation}
        \sigma_{F/\phi}\nabla^2\phi(x)\le \nabla^2 F(x)\le L_{F/\phi}\nabla^2\phi(x),
\end{equation}
We consequently define $\kappa_{F/\phi}=\frac{L_{F/\phi}}{\sigma_{F/\phi}}$ the relative condition number of $F$ with respect to $\phi$.
\end{defn}
Taking $\phi=\frac{1}{2}\NRM{.}^2$ gives $D_\phi=\frac{1}{2}\NRM{.}^2$ and thus yields classical smoothness and strong-convexity definitions. The idea of preconditioning is then to replace $\frac{1}{2\eta_t} \NRM{x-x_t}^2$ in \eqref{eq:proximal_gradient_descent} by $D_\phi(x,y)$ for a \emph{convenient} function $\phi$ which the server has access to, leading to:
\begin{equation}
    x_{t+1}\in \argmin_{x\in \R^d} \left \{ \langle \nabla F(x_t),x \rangle + \psi(x) +\frac{1}{\eta_t} D_\phi(x,x_t) \right \}.\label{eq:precond_proximal_gradient_descent}
\end{equation}
With $\eta_t=1/L_{F/\phi}$, the sequence generated by \eqref{eq:precond_proximal_gradient_descent} satisfies:
\begin{equation}
    \Phi(x_t)-\Phi(x^*)\le L_{F/\phi}(1-\kappa_{F/\phi}^{-1})^t.\label{eq:rate_kappa}
\end{equation}
Hence, the effectiveness of preconditioning hinges on how smaller $\kappa_{F/\phi}$ is compared to $\kappa_F$. Next subsection presents how our large deviation bound of Hessians (Proposition \ref{prop:deviation_hess}) comes into place. The better $\phi$ approximates $F$, the smaller $\kappa_{F/\phi}$ and the more efficient each iteration of \eqref{eq:precond_proximal_gradient_descent} is.

\subsection{Main Results in Statistical Preconditioning \label{section:statistical_precond_general}}

We furthermore assume that $F(x)=f(x)+\frac{\lambda}{2}\NRM{x}^2$ where $f$ verifies Assumption \ref{hyp:erm} and~$\lambda>0$. Assume that the server has access to an \emph{i.i.d.}~sample $\Tilde{a}_1,...,\Tilde{a}_N$ of the same law as the $a_{j}$'s and to functions $\Tilde{\ell}_1,...,\Tilde{\ell}_N$ such that $\Tilde{\ell}_i''=\ell''$. Define $\Tilde{f}(x)=\frac{\lambda}{2}\NRM{x}^2+\frac{1}{N}\sum_{i=1}^N\Tilde{\ell}_i(a_i^\top x)$. The preconditioner $\phi$ is chosen as, for some $\mu>0$:
\begin{equation}
    \phi(x)=\frac{\lambda}{2}\NRM{x}^2+\frac{1}{N}\sum_{i=1}^N\Tilde{\ell}_i(\Tilde{a}_i^\top x) +\frac{\mu}{2}\NRM{x}^2,
\end{equation}
Parameter $\mu>0$ is chosen such that, with high probability:
\begin{equation}
    \forall x \in \Dom_\psi, \NRM{\nabla^2 \Tilde{f} (x)-\nabla^2F(x)}_\op \le \mu.
\end{equation}
For such a $\mu>0$, we have: $L_{F/\phi}\le 1$, $\sigma_{F/\phi}\ge (1+2\mu/\lambda)^{-1}$ and $\kappa_{F/\phi}\le 1+\frac{2\mu}{\lambda}$. Recall that for $t=0,1,2,...$, we have $\NRM{x_t-x^*}^2\le C(1-\kappa_{F/\phi})^t$.
\begin{prop}[Statistical Preconditioning: Non-Isotropic Results]\label{prop:mu_precond}Assume that for all $x\in \Dom_\psi$, $\NRM{x}\le R$. 
Under Assumption \ref{hyp:erm}, with probability $1-\delta$, we have:
\begin{align*}
    \sup_{\NRM{x}\le R}\NRM{\nabla^2 \Tilde{f} (x) -\nabla^2 F(x)} &\le  CR \sigma_1^3\NRM{\ell''}_{\Lip}\left(\frac{(\deff(3)\ln(d)+\ln(1/\delta))\sqrt{\deff(1)+\ln(n/\delta)}}{n}\right.\\
    &+\left.\frac{\sqrt{\ln(1/\delta)}+\sqrt{\deff(1)\ln(d)}}{\sqrt{n}}\right).
\end{align*}
If $\mu$ is taken as this upper bound, then we control the rate of convergence in \eqref{eq:rate_kappa} with:
\begin{equation}
    \kappa_{F/\phi}=1+\Tilde{O}\left\{\frac{R\sigma_1^3\NRM{\ell''}_{\Lip}}{\lambda}\max\left(\frac{ \sqrt{\deff(1)}}{\sqrt{n}},\frac{\sqrt{\deff(1)}\deff(3)}{n}\right)\right\},
\end{equation}
where $\Tilde{O}$ hides logarithmic factors in $d,n$ and $\delta^{-1}$.
\end{prop}
Contrast this with known results:
\begin{remark}[Statistical Preconditioning: Isotropic Results] Still under Assumption \ref{hyp:erm},
\cite{hendrikx2020statistically} obtained:
\begin{equation}
    \kappa_{F/\phi}=1+\Tilde{O}\left\{\frac{R\sigma_1^3\NRM{\ell''}_{\Lip}}{\lambda}\max\left(\sqrt{\frac{ d}{n}},\frac{d^{3/2}}{n}\right)\right\}.
\end{equation}
\end{remark}
The only parameter required is an upper-bound on $\deff(3)$ and $\deff(1)$ in order to tune $\mu$. Simply knowing that data are distributed according to a highly non-isotropic subgaussian law can thus improve the efficiency of statistical preconditioning, by decreasing drastically estimates of $\kappa_{F/\phi}$ and the number of samples required in the preconditioning function.

\section{Non-Isotropic Randomized Smoothing\label{section:smooth}}
\subsection{General Considerations on the Randomized Smoothing Technique}

Consider an objective function $f:\R^d\to \R$ and a known convex regularizer $\psi$. $f$ is assumed to be convex and $L$-Lispchitz for some $L>0$. We assume that $\Dom_\psi \subset \cB(0,R)$. The following minimization problem:
\begin{equation}\label{eq:pbm_min_random_smooth}
    \min_{x\in\Dom_\psi} \Phi(x):= f(x)+\psi(x)
\end{equation}
is potentially hard as $f$ is not necessarily smooth. Moreover, $f$ is assumed to be of the form:
\begin{equation}
    \forall x\in \R^d, f(x)=\E_a[F(x,a)],\label{eq:f_general_smooting}
\end{equation}
for some random variable $a$ and $F$ a convex function, Lipschitz in its first variable. The second difficulty is thus that $f$ may not be directly computable, and a stochastic framework is required.\\

\noindent \textbf{Principle of the randomized smoothing technique and description of the algorithm:} in order to both use acceleration techniques and stochasticity of the gradients, the objective function $f$ is approximated by a smoothed version $f^\gamma$, where $\gamma>0$ is a parameter of the algorithm:
\begin{equation}
    \forall x\in\R^d, f^\gamma(x)=\E_Z[f(x+\gamma Z)]
\end{equation}
where $Z$ is a random variable, following a \emph{smoothing ditribution} $\mu$. \cite{scaman2018optimal} consider isotropic gaussians ($\mu=\cN(0,I_d)$), while \cite{duchi2012randomized} consider more general smoothing distributions (encompassing uniform distributions on the euclidean ball or on the $\ell^1$-ball). 
The algorithm then:
\begin{enumerate}
    \item Draws $Z_{1,t},...,Z_{m,t}$ \emph{i.i.d.}~random variables according to the smoothing distribution $\mu$, for $m$ a fixed integer.
    \item Queries the oracle at the m points $y_t+u_t Z_{i,t},i=1,...,m$, yielding stochastic gradients $g_{i,t}\in \partial F(y_t+u_t Z_{i,t},a_{i,t})$, where $y_t$ is the query point.
    \item Computes the average $g_t=\frac{1}{m} \sum_{i=1}^m g_{i,t}$.
    \item Uses this estimated gradient to perform an accelerated stochastic and proximal gradient step.
\end{enumerate}
For brevity, precise formulations of the algorithm and in particular of that last point are deferred to Appendix \ref{app:smooth}.

\subsection{Isotropic Randomized Smoothing}

We restrict ourselves to gaussian smoothing distributions $\mu$. In the isotropic case $\mu=\cN(0,I_d)$ considered by \cite{duchi2012randomized,scaman2018optimal}, the following crucial property holds, leading to a trade-off between precision and the smoothness parameter of $f^\gamma$.
\begin{prop}[Properties of Isotropic Gaussian Smoothing]\label{prop:iso_smooth} Let $\gamma>0$ and assume that $\mu=\cN(0,I_d)$. Recall that $f^\gamma(x)=\E_{Z\sim\cN(0,I_d)}[f(x+\gamma Z)]$ and $f$ is $L$-Lipschitz. We have:
\begin{equation}
    \forall x \in \R^d, f(x)\le f^\gamma(x) \le f(x)+\gamma L \sqrt{d},
\end{equation}
and $f^\gamma$ is $\frac{L}{\gamma}$-smooth. In order to reach an $\eps>0$ precision, one can take $\gamma=\frac{\eps}{L\sqrt{d}}$, for which $f^\gamma$ is then $\frac{L^2\sqrt{d}}{\eps}$-smooth.
\end{prop}
\begin{prop}[Convergence Guarantees with Isotropic Smoothing]\label{prop:iso_smoothing_conv} Take $\mu=\mathcal{N}(0,I_d)$ for the smoothing distribution. For a smoothing parameter $\gamma=Rd^{-1/4}$ and varying stepsizes in the accelerated gradient descent (Appendix \ref{app:smooth}), we have:
\begin{equation}
    \E[f(x_T)+\psi(x_T)-f(x^*)-\psi(x^*)] \le \frac{10 LRd^{1/4}}{T}+\frac{5LR}{\sqrt{Tm}},\label{eq:smooth_iso}
\end{equation}
and this $d^{1/4}$ factor cannot be improved: there exist objective functions $f$ and dimension-free constants such that we have an effective $d^{1/4}$ dependency in the global dimension.
\end{prop}
\noindent For $m$ big enough, the dominant term is $O(LRd^{1/4}/T)$, a dimensional dependency that cannot be alleviated \citep{nemirovsky1985,duchi2012randomized,bubeck2019complexity}.

\subsection{Non-Isotropic Randomized Smoothing}

In order to improve over Proposition \ref{prop:iso_smoothing_conv}, as it is optimal on the class of Lipschitz functions of the form \eqref{eq:f_general_smooting}, more assumptions are required in order to take advantage of an enventual underlying small effective dimension. We restrict ourselves to empirical measures of subgaussian random variables for $\nu$ in \eqref{eq:f_general_smooting} and to an empirical risk assumption for linear models such as in Assumption~\ref{hyp:erm}. We will hence assume that:
\begin{equation}
f(x)=\frac{1}{n}\sum_{i=1}^n\ell_i(a_i^\top x),
\end{equation}
for convex functions $\ell_i$, and $\Sigma$-subgaussian random variables~$a_i$. We furthermore assume that each $\ell_i$ is $L_\ell$-Lispchitz. Our interest in empirical measures lies in the fact that in practice one does not have access to an infinite number of samples. Our assumptions encompass non-smooth losses, such as $\ell_i(x)=\max(0,a_i^\top x- b_i)$.
%\begin{hyp}\label{hyp:smooth_duchi_non_iso}
%Random variable $a\sim \nu$ is $\Sigma$-subgaussian.  Denoting $\nu_n=\frac{1}{n}\sum_{i=1}^n\delta_{a_i}$ %an empirical measure with $a_1,...,a_n$ \emph{i.i.d.}~$\sim \nu$, there exists a $L_\ell$-Lipschtiz %function $\ell$ on $\R$ such that:
%\begin{equation}
%    \forall x \in \R^d, f(x)=\E_{\Tilde{a}\sim \nu_n}[\ell(\Tilde{a}^\top %x)]=\frac{1}{n}\sum_{i=1}^n\ell(a_i^\top x).
%\end{equation}
%\end{hyp}
%\mathieu{Peut être partir de cette forme là dès le début? Permet d'avoir des guaranties très similaires mais moins général pour la section précédente.}
As in Proposition \ref{prop:mu_precond}, one can hope to replace the $d^{1/4}$ factor in Proposition \ref{prop:iso_smoothing_conv} by an effective dimension dependent factor. A non-isotropic analog of Proposition \ref{prop:iso_smooth} for a smoothing distribution $\mu$ of the form $\cN(0,\Sigma')$ is required. It is quite intuitive to conjecture that adapting the smoothing distribution to the distribution of the data should indeed improve the efficiency of the algorithm. An analysis in the appendix shows that an optimal $\Sigma'$ is $\sqrt{\Sigma}$, hence the following proposition.
\begin{prop}[Properties of Non-Isotropic Gaussian Smoothing] \label{prop:smooth_prop_non_iso}  Set $\mu=\cN(0,\sqrt{\Sigma})$, $\gamma>0$. We have, with probability $1-\delta$:
\begin{equation}\begin{array}{ccc}\label{eq:smooth_non_iso}
     \forall x \in \R^d,  f^\gamma(x) &\le& f(x)+\gamma L_\ell\sqrt{\sigma_1^3(\deff(1)+\ln(n\delta^{-1}))\deff(2)}\\
     \NRM{\nabla f^\gamma}_{\Lip} & \le &\frac{L_{\ell}\sigma_1^{1/2}\deff(2)^{1/2}}{\gamma\deff(1)}\left(1+C\sqrt{\frac{\deff(1)\ln(d)+\ln(\delta^{-1})}{N}}\right).
\end{array}
\end{equation}
In order to reach an $\eps>0$ precision, one can take $\gamma=\frac{\eps}{L_\ell\sqrt{\sigma_1^3\deff(1)\deff(2)}}$, for which $f^\gamma$ is then $\frac{\sigma_1^2L_\ell^2\deff(2)}{\eps \sqrt{\deff(1)}}(1+C\sqrt{\frac{\deff(1)\ln(d)+\ln(\delta^{-1})}{N}})$-smooth with probability $1-\delta$.
\end{prop}
\begin{prop}[Convergence Guarantees with Non-Isotropic Smoothing] \label{prop:non_iso_smooth_conv}Taking $\mu=\mathcal{N}(0,\sqrt{\Sigma})$, for time-varying stepsizes defined in Appendix \ref{app:smooth}, we have with probability $1-\delta$ conditionally on the random variables $a_i,\Tilde{a}_j$:
\begin{equation}
    \E[f(x_T)+\psi(x_T)-f(x^*)-\psi(x^*)] \le \Tilde{O}\left(\frac{ LR\sigma_1\sqrt{\deff(2)/\sqrt{\deff(1)}}}{T}\right)+\frac{5L_0R}{\sqrt{Tm}},
\end{equation}
where $\Tilde{O}$ hides logarithmic factors in $d$ and $\delta^{-1}$.
\end{prop}
\noindent\eqref{eq:smooth_non_iso} corresponds to \eqref{eq:smooth_iso}, with $d$ replaced by $\frac{\deff(2)^2}{\deff(1)}$. Taking advantage of the underlying geometric repartition of the data thus yields better convergence guarantees, if we assume a more restrictive structure on the objective function. The knowledge of $\Sigma$ is here required to apply the previous considerations, whereas in the previous section only $\Tr(\Sigma)$ is needed. One may wonder to what extent our assumptions on $f$ could be generalized in order to obtain similar results.

%This emphasizes the importance of the effective dimension of the data. Furthermore, one may wonder to what extent these considerations could be generalized. More precisely, Assumption \ref{hyp:smooth_duchi_non_iso} implies that the contour lines of the objective function $f$ are distributed as ellispoids (that depend on matrix $\Sigma$). A more general structure than the one we considered may yield the same result, and could be an interesting problem to tackle. A starting point should be to try to model how an effective dimension could be involved in $f$ or $\nabla f$. 
%\mathieu{En fait ce serait surement dire que $f$ est smooth relativement à une métrique générée par la matrice $\Sigma$, mais pas sûr.}

\section{Conclusion}
%blablabla. Ouverture: autres manières de considérer des dimensions effectives plus petites (doubling, mesures spectrales,...). Ouverture dimension infinie. Généralisation de nos résultats à des fonctions plus générales (autres hypothèses, et forme plus générale de la variable aléatoire que l'on borne). Bornes de concentration plus tight (deff(1) au lieu de deff(r)). Autres problèmes de dimension (matrices aléatoires, ...). Problème du test de la sous-gaussianité.\\

Achieving \emph{effective dimension}-dependent bounds thus yields several applications, and we believe many others than the ones we studied exist. Broadening the set of applications could be achieved by: considering more general random variables, other models of effective dimension such as spectral dimension \citep{Durhuus:2009zz} or doubling dimension \citep{karbasi2012comparisonbased}, and infinite dimension $d$ but finite effective dimension such as in Appendix \ref{app:ellips_infinite} in order to take into account functional spaces for instance. 
Also, efficient methods for testing $\Sigma$-subgaussianity do not seem to exist, which should be an interesting problem to tackle.

\paragraph{Acknowledgements} The authors would like to thank Nikita Zhivotovskiy for very useful comments regarding Section \ref{section:tensors} and for pointing out relevant references.

\nocite{*}
\bibstyle{apalike} 
\bibliography{biblio}
\appendix
\renewcommand*\contentsname{Summary of the Article and of the Appendix}

\tableofcontents

\section{Covering Ellipsoids with Balls \label{section:appedix_ellipsoids}}

\subsection{Proof of Theorem \ref{thm:ellipsoid_non_asympt}}
\cite{Dumer2004OnCO} prove the asymptotic version of our result. We use their method in order to prove Theorem \ref{thm:ellipsoid_non_asympt} in what follows. 

\begin{proof} The proof involves three steps. In the first one, we cover ellipsoid $E_b$ by direct products of balls of lesser dimensions. Then, in Step 2, we derive a general upper bound. Finally, tuning our parameters from the bound obtained in Step 2 leads to the desired result in Step 3. Steps 1 and 2 of the proof are directly inspired by \citet{Dumer2004OnCO}, while step 3 and the corollaries that follow are the non-asymptotic improvements over their results.
\\

\noindent \textbf{Step 1.} Let $t\in \N,t\le d$, $0=n_0<n_1<...<n_t=d$ and $I_i=\{n_{i-1}+1,...,n_i\}$ for $i=1,...,t$, in order to divide $[d]$ into $t$ subsets. Let $s_i=n_i-n_{i-1}$ for $i=1,...,t$. For some parameter $h\in (0,1)$ let the set of numbers:
\begin{equation*}
    H=\{ih, i=1,...,\left \lfloor h^{-1}\right \rfloor +1\}.
\end{equation*}
For any $w\in [0,1]$, let $\bar{w}$ be the closest point in $H$ exceeding $w$. Consider the following subset of~$H^t$:
\begin{equation*}
U=\{(u_1,...,u_t)\in H^t|\sum_{i=1}^tu_i\le 1+th\}
\end{equation*}
Let $u\in U$ be fixed. For $i=1,...,t$, consider the ball of dimension $s_i$:
\begin{equation*}
    B_i^u=\left\{ x \in \R^{I_i} | \sum_{j\in I_i} x_j^2\le \rho_i^2\right\}, \text{ where } \rho_i^2=u_i b_{n_{i-1}+1}^2.
\end{equation*}
Let $D_u$ be the direct product of all $t$ balls:
\begin{equation*}
    D_u=\prod_{i=1}^tB_i^u=\left\{ x\in \R^d| \sum_{j\in I_i}^d x_j^2/b_{n_{i-1}+1}^2 \le u_i, i=1,...,t\right\}.
\end{equation*}
We have:
\begin{equation*}
    E_b\subset \bigcup_{u\in U} D_u.
\end{equation*}
Indeed, for $x\in E_b$, let $w_i=\sum_{j\in I_i}x_j^2/b_j^2$  and take $u_i=\bar{w_i}$. First, $x_{I_i}\subset B_i$:
\begin{equation*}
    \sum_{j\in I_i} x_j^2/b_{n_{i-1}+1}^2 \le  \sum_{j\in I_i} x_j^2/b_j^2 = w_i\le u_i.
\end{equation*}
Moreover, $u\in U$:
\begin{equation*}
    \sum_{i=1}^t u_i \le \sum_{i=1}^t w_i+h \le 1 + th.
\end{equation*}
Hence, for any $x\in E_b$, there exists $u\in U$ such that $x\in D_u$.\\

\noindent \textbf{Step 2.} Given $D_u$ for some $u\in U$, denote $\cH_1(D_u)$ its unit entropy. We have:
\begin{equation*}
    \cH_1(D_u)=\inf_{e\in ({R^+}^*)^t: \sum_i e_i^2\le 1} \sum_{i=1}^t\cH_{e_i}(B_i^u),
\end{equation*}
where $\cH_{e_i}(B_i^u)$ is the $e_i$-entropy of $B_i^u$. As $E_b\subset \bigcup_{u\in U} D_u$, we have that:
\begin{equation*}
    \cH_1(E_b)\le \ln(|U|)+\sup_{u\in U}\left( \inf_{e\in ({R^+}^*)^t: \sum_i e_i^2\le 1} \sum_{i=1}^t\cH_{e_i}(B_i^u) \right ).
\end{equation*}
We have:
\begin{equation*}
    |U|=\binom{t+\lfloor h^{-1} \rfloor}{t}:=\cN(t,h).
\end{equation*}
In order to estimate quantities such as $\cH_{e_i}(B_i^u)$, we will need results on the $\eps$-entropies of balls that directly come from \cite{rogers_1963}:
\begin{lemma}
For any dimension $d>0$, any ball $B_\rho$ of radius $\rho>0$ has a unit entropy $\cH_1(B_\rho)$ upper-bounded by:
\begin{equation}
    \cH_1(B_\rho)\le n\ln(\rho)+c\ln(n+1),
\end{equation}
for some universal constant $c>0$.
\end{lemma}
\end{proof}
\noindent Using this, we obtain:
\begin{equation*}
    \cH_1(E_b)\le \ln(\cN(t,h))+\sup_{u\in U}\left( \inf_{e\in ({R^+}^*)^t: \sum_i e_i^2\le 1} \sum_{i=1}^t s_i\ln(\rho_i/e_i) + c\ln(s_i+1) \right ).
\end{equation*}
As $\rho_i^2=u_ib_{n_{i-1}+1}^2$, for fixed $u\in U$ we have:
\begin{align*}
    \inf_{e\in ({R^+}^*)^t: \sum_i e_i^2\le 1} \sum_{i=1}^t s_i\ln(\rho_i/e_i) & = \inf_{e\in ({R^+}^*)^t: \sum_i \eps_i\le 1} \frac{1}{2}\sum_{i=1}^t s_i\ln(u_ib_{n_{i-1}+1}^2/\eps_i)\\
    & \le \frac{1}{2}\sum_{i=1}^ts_i\ln(u_i) + \sum_{i=1}^t s_i \ln^+(b_i)\\
    & \le \ln(\gamma)\sum_{i=1}^ts_i + \sum_{i=1}^t s_i\ln^+(b_i),
\end{align*}
where we note $\gamma=\sqrt{1+th}$. Now consider $\hat{b}\in \R^d$ the vector with coefficients:
\begin{equation*}
    \hat{b}_j=b_{n_{i-1}+1}, j\in I_i,i=1,...,t,
\end{equation*}
such that:
\begin{equation*}
    \sum_{i=1}^t s_i\ln^+(b_i)= K_{\hat{b}}.
\end{equation*}
Furthermore, comparing $K_b$ and $K_{\hat{b}}$:
\begin{align*}
    \sum_{i=1}^t s_i\ln^+(b_i) &\le \sum_{j=1}^d \ln^+(b_j) + \sum_{i=1}^{t-1} \sum_{j\in I_i} \ln(b_{n_{i-1}+1}/ b_i)\\
    &\le K_b + \sum_{i=1}^{t-1} (s_i-1) \ln(b_{n_{i-1}+1}/b_{n_i}).
\end{align*}
The sum above ends at $t-1$ by definition of $m$ and of the interval $I_t$.
We hence obtain the following general upper-bound on $\cH(E_b)$, concluding Step 2:
\begin{equation}
    \cH(E_b)\le K_b + \ln(\cN(t,s)) + \sum_{i=1}^t (s_i-1) \ln(b_{n_{i-1}+1}/b_{n_i}) + n\ln(\gamma) + c\sum_{i=1}^t\ln(s_i+1).\label{eq:step_3_ell}
\end{equation}
\\

\noindent \textbf{Step 3.} To provide the desired result, we tune $h,n_1,...,n_t,t,s_1,...,s_t$ in the following way. Let $h=1/d$. For simplicity, we denote $m=m_b=\sum_{i: b_i>1} 1$. We choose $s_t=d-m$ and for $i=2,...,t-1$, we set $s_i=s$ for some $s\in \N^*$ to determine. We have $s_1\le s$, and $t\le 1 + \lceil m/s \rceil$. Let us bound the terms appearing in \eqref{eq:step_3_ell} from left to right. 
\begin{align*}
    \ln(\cN(t,s))\le \ln( (e(1+t^{-1}h^{-1}))^t) \le t (1+\ln(1+d)).
\end{align*}
Then, since $s_1\le s_2=...=s_{t-1}=s$, and by definition of $m$:
\begin{align*}
    \sum_{i=1}^{t-1} (s_i-1) \ln(b_{n_{i-1}+1}/b_{n_i})= (s-1)\ln(b_1/b_m)\le (s-1) \ln(b_1).
\end{align*}
We chose $h=1/n$ such that, using $\gamma=\sqrt{1+th}$:
\begin{align*}
    n\ln(\gamma)=\frac{n}{2}\ln(1+t/n)\le t/2.
\end{align*}
Finally:
\begin{align*}
    \sum_{i=1}^t\ln(s_i+1) = (t-1)\ln(s+1)+\ln(d-m+1).
\end{align*}
Combining these inequalities leads to:
\begin{equation*}
    \cH_1(E_b)\le K_b + C\left( t\ln(d) + (s-1) \ln(b_1) \right).
\end{equation*}
As $t\le 1+\lceil m/s \rceil$:
\begin{equation*}
    \cH_1(E_b)\le K_b + C\left( (1+\lceil m/s \rceil)\ln(d) + (s-1) \ln(b_1) \right).
\end{equation*}
We now need to tune $s$. We take $s$ of the form:
\begin{equation*}
    s=\left \lceil \frac{m\ln(d)}{\eta} \right \rceil,
\end{equation*}
for some $\eta>0$, leading to:
\begin{equation*}
    \cH_1(E_b)\le K_b + C\left( (1+\lceil \eta  / \ln(d) \rceil)\ln(d) + (\left \lceil \frac{m\ln(d)}{\eta } \right \rceil-1) \ln(b_1) \right).
\end{equation*}
Using $\lceil x\rceil \le 1+2x$ for any $x\ge 0$:
\begin{equation*}
    \cH_1(E_b)\le K_b + C\left( 2\eta  +  2\frac{m\ln(d)}{\eta }  \ln(b_1) + 2\ln(d)\right).
\end{equation*}
Optimizing and taking $\eta= \sqrt{m\ln(d)\ln(b_1)}$ gives:
\begin{equation*}
    \cH_1(E_b)\le K_b + C\left( 4\sqrt{m\ln(d)\ln(b_1)} + 2\ln(d)\right),
\end{equation*}
concluding our proof.
\subsection{Proof of Corollary \ref{cor1}}

We start by proving Corollay \ref{cor1}.
Let $\eps>0$. Consider the ellipsoid $E_b$, where $b_i=\eps^{-1}\sigma_i/\sigma_1$, $i\in[d]$, and a covering $C(E_b)$ of $E_b$ by unit Euclidean balls. Theorem \ref{thm:ellipsoid_non_asympt} gives us an upper bound on the minimal size of such coverings. 
Let $S=\hbox{Diag}(\eps^{-1}\sigma_i/\sigma_1)_{i\in[d]}$. We then define $\cN_\eps= S^{-1} C(E_b)$. By definition of $E_b$, $x\in \cS_1$ if and only if $Sx\in E_b$, so that $\cN_\eps$ consists of vectors in $\cS_1$. Moreover, for each $y\in \cS_1$, by definition of $C(E_b)$, there exists $x\in \cN_\eps$ such that $\|Sy-Sx\|^2\le 1$. This is precisely the condition required.\\

We now derive an upper bound on $\cH_\eps$ in terms of $\deff(r)$. Let $m_\eps$ be given, so that:
$$
\eps^{-1} \sigma_{m_\eps}/\sigma_1>1\ge \eps^{-1}\sigma_{m_\eps+1}/\sigma_1.
$$
Given $\deff(r)$, the value of $\sigma_{m_\eps}$ is maximized by taking the values of $\sigma_2^2/\sigma_1^2,\ldots,\sigma_{m_\eps}^2/\sigma_1^2$ all equal to $(\deff(r)-1)/(m_\eps-1)$. We thus find that necessarily,
$$
\eps^{-\frac{2}{r}}(\deff(r)-1)\ge m_\eps-1.
$$
We also have the trivial bound $m_\eps-1\le d-1$.  Next, we note that, for fixed $m_\eps$, the value of $\sum_{i=1}^{m_\eps}\ln(\eps^{-1}\sigma_i/\sigma_1)$ is maximized again, using concavity of $\ln$, by taking $\sigma_2^{2/r}/\sigma_1^{2/r},\ldots,\sigma_{m_\eps}^{2/r}/\sigma_1^{2/r}$ all equal to $(\deff(r)-1)/(m_\eps-1)$.  This then evaluates to:
$$
\sum_{i=1}^{m_\eps}\ln(\eps^{-1}\sigma_i/\sigma_1)=\ln(\eps^{-1})+\frac{m_\eps-1}{2/r}\ln(\eps^{-\frac{2}{r}}(\deff(r)-1)/(m_\eps-1))
$$
We then use the fact that $x\to x\ln(A/x)$ is increasing over $[0,A/e]$. Here, $x$ plays the role of $m_\eps-1$, and $A$ the role of $\eps^{-\frac{2}{r}}(\deff(r)-1)$. We end up with:
$$
\sum_{i=1}^{m_\eps}\ln(\eps^{-1}\sigma_i/\sigma_1)\le \ln(\eps^{-1})+\frac{\min(d-1,\eps^{-\frac{2}{r}}(\deff(r)-1)/e)}{2/r}\ln(\max(e,\eps^{-\frac{2}{r}}(\deff(r)-1)/(d-1))).
$$

\subsection{Ellipsoids in Infinite Dimension\label{app:ellips_infinite}}

We here present results on the unit entropy of ellipsoids in infinite dimension, which we believe to be of interest. More precisely, consider the space $V=\ell^2(\R)=\{x\in\R^{\N^*}|\sum_{i\ge 1}x_i^2<\infty\}$ with the classical euclidean topology. We have $\NRM{x}^2=\sum_{i\ge 1}x_i^2$ for $x\in V$. Note however that what follows can naturally be extended to any separable Hilbert space.
\begin{defn}
For $b\in V$ such that $b_1\ge b_2\ge...>0$, we define the ellipsoid $E_b$ as:
\begin{equation}
    E_b=\left \{ x\in V : \sum_{i\ge 1} \frac{x_i^2}{b_i^2}\le 1 \right\}.
\end{equation}
\end{defn}
We then define the $\eps$-entropy and unit-entropy of such an ellipsoid as in the finite-dimension case. 
\begin{thm}[Unit Entropy of Ellipsoids in Infinite Dimension] \label{thm:ell_infinite} Let $E_b\subset V$ an ellipsoid. Define the following quantities:
\begin{equation*}
\begin{array}{ccc}
    K_b & = & \sum_{i\ge 1}\ln^+(b_i),\\
    m_b & = & \sum_{i\ge 1} \II_{b_i\ge 1/2}\\
    M_b & = & \inf\{ n\ge 1 : \sum_{i\ge n+1} b_i^2 \le 1/2 \}
\end{array}
\end{equation*}
Then, we have for some universal constant $c>0$:
\begin{equation}
    \cH_1(E_b)\le K_b +c\left(\sqrt{\ln^+(b_1)\ln(M_b)m_b}+\ln(M_b)\right).
\end{equation}
\end{thm}
The proof follows the same steps as the one in finite dimension, replacing the global dimension in Step 1 by $M_b$. The interest of ellipsoids in infinite dimension lies in the appearance of another notion of dimension than the one we studied: the power-law norm decay of a vector $v\in V$.
\begin{defn}[Power-Law Norm Decay] A vector $x\in V$ is of power-law norm decay $d>0$ if:
\begin{equation*}
    \sum_{i\ge n+1}x_i^2 = \mathcal{O}\left(n^{-2/d}\right).
\end{equation*}
An ellipsoid $E_b$ is said of power-law norm decay $d$ if $b$ is of power-law norm decay $d$.
\end{defn}
The power-law norm decay $d$ of a vector $\lambda\in V$ is closely related to the spectral dimension of infinite graphs or operators on Hilbert spaces: $(\lambda)_{i\in\N^*}$ usually corresponds to the eigenvalues of the Laplacian of the graph in the first case, or to the eigenvalues of the operator in the second case.  The following corollary illustrates how this notion is relevant.
\begin{cor}[$\eps$-Entropy and Power-Law Norm Decay] Let $E_b\subset V$ be an ellipsoid of power-law norm decay $d>0$. Then, when $\eps\to 0$, we have:
\begin{equation}
    \cH_\eps(E_b) \le d \ln(\eps^{-1})^2 (1 + o(1)).
\end{equation}
\end{cor}
\begin{proof}
We have for any ellipsoid $E_b$ of power-law norm decay $d$: \begin{equation}
    m_{b/\eps}\le M_{b/\eps} = \mathcal{O}(\eps^{-d/2}).\label{eq:m_b_M_b}
\end{equation}
Remark that $\cH_\eps(E_b)=\cH_1(E_{b/\eps})$. Then, using Theorem \ref{thm:ell_infinite}, and the previous consideration \eqref{eq:m_b_M_b}, we obtain our result.
\end{proof}

This needs to be put in light with the $\eps$-entropy of the unit ball in dimension $d<\infty$, that behaves as $d\ln(\eps^{-1})$ when $\eps\to0$. Despite the presence of $\ln(\eps^{-1})^2$ instead of $\ln(\eps^{-1})$, we have a linearity in this expression in terms of $d$. 

\section{Proof of Theorems \ref{thm:chaining_not_centered} and \ref{thm:chaining_centered} and General Considerations on these Large Deviation Bounds}
\subsection{Proof of Theorem \ref{thm:chaining_not_centered}: Bound Without Centering}

In order to have lighter notations, we write $\deff'=\deff(r)$ in what follows.
For all $j\ge 0$, let $\cN_j$ be a covering of $\cS_1$ satisfying the properties of Corollary \ref{cor1} for $\eps_j=2^{-j}$.
For all $x\in\cS_1$, let $\Pi_j x$ be some point in $\cN_j$ such that \eqref{eq:cor1} holds. By convention we take $\Pi_0 x=0$.
Then for all $(x_1,\ldots,x_r)\in\cS^r$, using the chaining approach, we write
\begin{align*}
\frac{1}{n}\sum_{i\in[n]}\prod_{k\in[r]}f_k(a_i^\top x_k)&=\sum_{j\ge 0}\frac{1}{n}\sum_{i\in[n]}\left\{
\prod_{k\in[r]}f_k(a_i^\top \Pi_{j+1} x_k)-\prod_{k\in[r]}f_k(a_i^\top \Pi_{j}x_k)\right\}\\
&=\sum_{j\ge 0}\sum_{k\in[r]}\frac{1}{n}\sum_{i\in[n]}[f_k(a_i^\top\Pi_{j+1}x_k)-f_k(a_i^\top\Pi_j x_k)]\\
&\times\left \{\prod_{\ell=1}^{k-1}f_{\ell}(a_i^\top\Pi_{j+1}x_{\ell}) \times\prod_{\ell=k+1}^r f_\ell(a_i^\top \Pi_j x_\ell)\right\}.
\end{align*}
%For fixed $j\ge 0$ and $r\in[k]$, there are at most $|\cN_j|^r |\cN_{j+1}|^r$ possible choices for the vectors $\Pi_j x_k$, $\Pi_{j+1}x_k$ appearing in the corresponding term. 
Let $j\ge 0$ and $k\in[r]$ be fixed.
Consider a term of the form
$$
Z=\frac{1}{n}\sum_{i\in[n]}Z_i,
$$
with 
\begin{equation}\label{eq:z_i}
Z_i=\prod_{\ell=1}^{k-1} f_\ell(a_i^\top u_{\ell})[f_k(a_i^\top u_k)-f_k(a_i^\top v_k)]\prod_{\ell=k+1}^r f_\ell(a_i^\top v_\ell),
\end{equation}
where $u_\ell\in\cN_j$, $v_\ell\in\cN_{j+1}$, and $\|u_k-v_k\|_{\Sigma}\le \sigma_1\epsilon_j$, where we defined $\epsilon_j:=2^{-j+1}$. By the triangle inequality, for all $x_\ell\in \cS_1$, letting $u_\ell=\Pi_j x_\ell$ and $v_\ell=\Pi_{j+1}x_\ell$, these assumptions are satisfied. Note also that $\|u_\ell\|_\Sigma$ and $\|v_\ell\|_\Sigma$ are upper-bounded by  $\sigma_1$.

\noindent Clearly,  $|Z_i|\le 2B$. Also, 
$$
\begin{array}{ll}
|Z_i|&\le (B/B_k)|a_i^\top(u_k-v_k)|\\
&\le (B/B_k)\|a_i\|_{\Sigma^{-1}} \|u_k-v_k\|_{\Sigma}.
\end{array}
$$
We introduce the new parameter $R_{\Sigma^{-1}}$, that is an upper bound on the norms $\|a_i\|_{\Sigma^{-1}}$.
Note that, for the Gaussian case where $\Sigma$ is the covariance matrix of the Gaussian vector $a_i$, the natural scaling assumption is to take:
$$
R_{\Sigma^{-1}}=\Theta(\sqrt{d}).
$$
As we don't want any dependency on the overall dimension $d$, we will aim at making this quantity disappear. We then introduce the notation
$$
P_{j,k}:=\min(2B/\epsilon_j,(B/B_k)\sigma_1 R_{\Sigma^{-1}}).
$$
For each $Z_i$ and $t>0$, using that $\NRM{u_k-v_k}_\Sigma\leq \eps_j$, we then have:
$$
\begin{array}{lll}
\dP(Z_i\ge\epsilon_j\sigma_1^rt)&\le& \dP(|f_\ell(a_i^\top u_\ell)|\ge \sigma_1 t^{1/r}\hbox{ for some }\ell<k,\\
&&\;\hbox{ or }|f_k(a_i^\top u_k)-f_k(a_i^\top v_k)|\ge \sigma_1 \epsilon_j t^{1/r},\\
&&\;\hbox{ or }|f_\ell(a_i^\top v_\ell)|\ge \sigma_1 t^{1/r}\hbox{ for some }\ell>k)
\\
&\le & 2r e^{-t^{2/r}/2}\hbox{ if } t\le \sigma_1^{-r}P_{j,k}\\
& =& 0 \hbox{ if }t> \sigma_1^{-r}P_{j,k},
\end{array}
$$
%\textbf{Following notation useless. }We will also make use of notation $j^*(k):=\lfloor %\ln_2(\sigma_1R_{\Sigma^{-1}}/B_k)\rfloor$, so that
%$$
%j\le j^*(k)\Rightarrow P_{j,k}=2B/\epsilon_j,\; j>j^*\Rightarrow %P_{j,k}=\sigma_1(B/B_k)R_{\Sigma^{-1}}.
%$$
We note $P_j=\frac{2B}{\eps_j}\ge P_{j,k}$. The previous bounds on the tail of $Z_i$'s distribution allow to bound exponential moments of $Z_i$. Fix some $\theta>0$. Then
$$
\begin{array}{lll}
\dE e^{(\theta/n)\sigma_1^{-r}Z_i/\epsilon_j}&\le&1+\frac{\theta}{n}\int_0^{\infty} e^{(\theta/n)y}\dP(\sigma_1^{-r}Z_i/\epsilon_j\ge y)dy\\
&\le & 1+\frac{\theta}{n}2r \int_0^{\sigma_1^{-r}P_{j,k}} e^{(\theta/n)y} e^{-y^{2/r}/2}dy.
\end{array}
$$
We now fix $\theta_{j,k}>0$ such that, for all $y\in[0,\sigma^{-r}P_{j,k}]$, one has:
$$
(\theta_{j,k}/n)y\le \frac{1}{4}y^{2/r},
$$
or equivalently, we assume (recall that $r\ge 2$):
\begin{equation}\label{eq:borne_theta}
\frac{\theta_{j,k}}{n}\le \frac{1}{4} (\sigma_1^{-r}P_{j,k})^{2/r-1}.
\end{equation}
This entails the bound:
$$
\dE e^{(\theta_{j,k}/n)\sigma_1^{-r}Z_i/\epsilon_j}\le 1+\frac{\theta_{j,k}}{n}2r \int_{0}^{\infty} e^{-y^{2/r}/4}dy=:1+\frac{\theta_{j,k}}{n}c_r,
$$
where we introduced notation $c_r:= 2r \int_{0}^{\infty} e^{-y^{2/r}/4}dy$.
Thus, for $\theta_{j,k}>0$ satisfying \eqref{eq:borne_theta}, one has:
\begin{equation}
\dE e^{\theta_{j,k}\sigma_1^{-r} \frac{1}{n}\sum_{i\in[n]}Z_i/\epsilon_j} \le (1+\theta_{j,k} c_r/n)^n\le e^{\theta_{j,k} c_r}.
\end{equation}
The number of possible choices for $u_\ell\in\cN_j$ and $v_\ell\in\cN_{j+1}$ involved in the definition of $Z_i$  is upper-bounded by
$$
|\cN_{j+1}|^{r+1}\le e^{(r+1)\cH_{j+1}},
$$
where $\cH_j$ is defined in \eqref{eq:cor2}.
Thus for any $t_{j,k}>0$, the probability that for some choice of $u_\ell$, $v_\ell$ in the corresponding nets, one has
$$
\frac{1}{n}\sum_{i\in[n]}Z_i \ge \epsilon_j \sigma^r t_{j,k}
$$
is upper bounded by:
$$
e^{(r+1)\cH_{j+1}} \inf_{\theta\in[0,\frac{n}{4}(\sigma^{-r}P_{j,k})^{2/r-1}]} e^{-\theta t_{j,k} +\theta c_r}.
$$
We now take $\theta_{j,k}=\frac{ n (\sigma_1^{-r}P_{j,k})^{2/r-1}}{4}$, and :
$$
t_{j,k}= c_r+\frac{(r+1)\cH_{j+1}+\lambda +2\ln(j+1)}{\theta_{j,k}}
$$
for some $\lambda\ge 0$.
This upper bound is then no more than $(1+j)^{-2}e^{-\lambda}$.
We now use a union bound over $j\ge 0$ and $k\in[r]$ to obtain:
$$
\dP(Y\ge \sigma^r\sum_{j\ge 0}\sum_{k\in[r]}\epsilon_j t_{j,k})\le e^{-\lambda}\; r\sum_{j\ge 0}(1+j)^{-2}=r\frac{\pi^2}{6}e^{-\lambda}.
$$

\noindent Let us bound the sum appearing in this probability. Fix some $k\in [r]$. 
\begin{align}\label{eq:sumgrosse}
    \sum_{j\ge 0}\epsilon_j t_{j,k}&=\sum_{j\ge 0}2^{-j+1}\left[c_r+\frac{(r+1)\cH_{j+1}+\lambda +2\ln(j+1)}{\theta_{j,k}}\right]
\end{align}
with $\theta_{j,k}\ge \frac{n}{4}\left(\sigma_1^{-r}P_j\right)^{2/r-1}=\left(\sigma_1^{-r}2B\right)^{2/r-1} 2^{-j+j\frac{2}{r}}$, for $P_{j}=2B/\eps_j$. Moreover, for $j^*=\frac{r}{2}\ln_2\left(e\frac{d-1}{\deff'-1}\right)$, we have:
\begin{align*}
    j\leq j^* &\implies\cH_j \le j\ln(2) +2^{j\frac{2}{r}}\frac{\deff'-1}{e}+c\left [\ln(d) +\sqrt{j\ln(d)m_j}\right],\\
    j>j^*&\implies \cH_j\le j\ln(2)+(d-1)\ln\left(2^{j\frac{2}{r}}\frac{\deff'-1}{d-1}\right)+c\left [\ln(d) +\sqrt{j\ln(d)m_j}\right],\\
    &\text{with } m_j\le 1+2^{j\frac{2}{r}}(\deff'-1).
\end{align*}
\noindent The \emph{easy} part of \eqref{eq:sumgrosse} to study:
\begin{align*}
    \sum_{j\ge 0}2^{-j+1}\left[c_r+\frac{\lambda+2\ln(j+1)}{\theta_{j,k}}\right]&\leq  \frac{A_r\lambda +B_r}{\frac{n}{4}\left(\sigma_1^{-r}B\right)^{2/r-1}}
\end{align*}
Now, let us bound $H:=\sum_{j>0}2^{-j}\cH_j/\eps_j^{1-2/r}=\sum_{j>0}2^{-j\frac{2}{r}}\cH_j$.
\begin{align*}
    \sum_{1\le j\le j^*}2^{-j\frac{2}{r}}\cH_j&\leq\sum_{1\le j\le j^*}2^{-j\frac{2}{r}}\left(j\ln(2) +2^{j\frac{2}{r}}\frac{\deff'-1}{e}+c\left [\ln(d) +\sqrt{j\ln(d)2^{j\frac{2}{r}}\deff'}\right]\right)\\
    &\leq C_r j^* \deff' + D_r \sqrt{\ln(d)\deff'}+E_r\ln(d).
\end{align*}
Using that $j^*$ is \emph{big} when $\deff'\ll d$, we bound the sum for $j>j^*$:
\begin{align*}
    \sum_{j>j^*}2^{-j\frac{2}{r}}\cH_j&\le  \sum_{j>j^*}2^{-j\frac{2}{r}}\left(j\ln(2)+(d-1)\ln\left(2^{j\frac{2}{r}}\frac{\deff'-1}{d-1}\right)+c\left [\ln(d) +\sqrt{j\ln(d)m_j}\right]\right)\\
    &\le F_r' \sum_{j>j^*}j2^{-j\frac{2}{r}}d + G_r'\sum_{j>j^*}2^{-j\frac{2}{r}}\sqrt{j2^{j\frac{2}{r}}\ln(d)}\\
    &\le F_r \deff'\ln\left(\frac{d}{\deff'}\right) + G_r,
\end{align*} where we used that $2^{j^*\frac{2}{r}}=e\frac{d-1}{\deff'-1}$ and $\sum_{j>k}j2^{-j\frac{2}{r}}\le C k2^{-k\frac{2}{r}}$. All in one, that leaves us with the following bound on \eqref{eq:sumgrosse}:
\begin{align*}
      \sum_{j\ge 0}\epsilon_j t_{j,k}&\le c_r\frac{A_r\lambda +B_r+C_r \deff' + D_r \sqrt{\ln(d)\deff'}+E_r\ln(d)+ F_r \deff'\ln\left(\frac{d}{\deff'}\right) + G_r}{\frac{n}{4}\left(\sigma_1^{-r}B\right)^{2/r-1}}.
\end{align*}
In a more synthetic formulation:
\begin{equation}
    \sum_{j\ge 0}\epsilon_j t_{j,k}\le C_r\left(1+\frac{ \deff'\ln(d)+\lambda}{\frac{n}{4}\left(\sigma_1^{-r}B\right)^{2/r-1}}\right).
\end{equation}
For some suitable constant $C_r$ the following holds, by summing previous considerations for $1\le k\le r$.
One has for all $\lambda>0$ that:
\begin{equation}
\dP\left(Y\ge \sigma_1^r C_r\left(1+\frac{\deff(r)\ln(d)+\lambda}{n}(\sigma_1^{-r} B)^{1-2/r}\right)\right)\le e^{-\lambda} \;r\frac{\pi^2}{6}.
\end{equation}

\subsection{Proof of Theorem \ref{thm:chaining_centered}: Bound With Centering}

We now look for bounds on:
\begin{equation}\label{eq:new_bound_2}
Y':=\sup_{x_1,\ldots,x_r\in \cS_1} \frac{1}{n}\sum_{i\in[n]}\left\{\prod_{k=1}^r f_k(a_i^\top x_k)-\dE\prod_{k=1}^rf_k(a_1^\top x_k)\right\}.
\end{equation}
Fixing $j\ge 0$, $k\in[r]$, we again consider the random variables $Z_i$ as previously defined in \eqref{eq:z_i}. Write:
$$
\dE e^{(\theta/n)\sigma_1^{-r}[Z_i-\dE Z_i]/\epsilon_j}=1+\left(\frac{\theta}{n}\right)^2 \dE \left[(\epsilon_j^{-1}\sigma_1^{-r}(Z_i-\dE Z_i))^2 F((\theta/n)(\epsilon_j^{-1}\sigma_1^{-r}(Z_i-\dE Z_i)))\right], 
$$
where :
$$
F(x):=x^{-2}[e^{x}-x-1]\le e^{|x|}.
$$
Thus using this bound and the inequality $xy\le x^2+y^2$:
$$
\dE e^{(\theta/n)\sigma_1^{-r}[Z_i-\dE Z_i]/\epsilon_j}\le 1+\left(\frac{\theta}{n}\right)^2\left[\dE((\epsilon_j^{-1}\sigma_1^{-r}(Z_i-\dE Z_i))^4+\dE e^{2(\theta/n)\sigma_1^{-r}|Z_i -\dE Z_i|/\epsilon_j}\right].
$$
By the sub-gaussian tail assumption, $\dE(\epsilon_j^{-1}\sigma_1^{-r}(Z_i-\dE Z_i))^4$ is bounded by a constant $\kappa_r$ dependent on $r$. By the same arguments as above, for:
$$
\frac{\theta}{n}\le \frac{(\sigma_1^{-r}P_{j,k})^{2/r-1}}{8},
$$
then $\dE e^{2(\theta/n)\sigma_1^{-r}|Z_i -\dE Z_i|/\epsilon_j}$ is also bounded by another constant $\kappa'_r$ dependent on $r$. Indeed, by the sub-gaussian tail assumption, $|\dE Z_i|\le \sigma_1^r \epsilon_j  s_r$ for some $r$-dependent constant, and we can then use the upper bound:
$$
\begin{array}{ll}
\dE e^{2\theta/n\sigma_1^{-r}|Z_i-\dE Z_i|/\epsilon_j}&\le e^{2(\theta/n)s_r}[1+\frac{\theta}{n}\int_0^{\infty}e^{2(\theta/n)y}[\dP(Z_i\ge y\sigma_1^r \epsilon_j)+\dP(-Z_i\ge y\sigma_1^r\epsilon_j)]dy]
\\
&\le e^{2(\theta/n)s_r}[1+\frac{\theta}{n}2r\int_0^{\sigma^{-r}P_{j,k}}e^{2(\theta/n)y-y^{2/r}/2}dy
].\end{array}
$$  

\noindent Thus the probability that for some choices of $u_\ell,v_{\ell},\; \ell\in[r]$ in the suitable $\epsilon$-nets, one has:
$$
\frac{1}{n}\sum_{i\in[n]}Z_i-\dE Z_i\ge \sigma_1^r \epsilon_j t_{j,k}
$$
is upper-bounded, for all $\theta\in[0,n(\sigma_1^{-r} P_{j,k})^{2/r-1}/8]$ by:
$$
\exp\left( (r+1)\cH_{j+1}- \theta t_{j,k} + \kappa''_r \theta^2/n\right),
$$
where we defined $\kappa''_r=\kappa_r+\kappa'_r$.
Let now $\theta_{j,k}=\min(n,n(\sigma_1^{-r}P_{j,k})^{2/r-1}/8)$, and :
$$
t_{j,k}=\kappa''_r \frac{\theta_{j,k}}{n}+\frac{1}{\theta_{j,k}}[(r+1)\cH_{j+1}+\lambda+2\ln(j+1)],
$$
where $\lambda>0$ is a free parameter.
We have, for $t_j=\sum_k t_{j,k}$, if all $\theta_{j,k}$ have the same value $\theta_j$:
$$
t_j\le c_r\left(\frac{\theta_j}{n}+\frac{(r+1)\cH_{j+1}+\lambda +2\ln(j+1)}{\theta_{j}}\right)
$$
We now take $\theta_{j}=\min\left(\frac{ n (\sigma_1^{-r}P_{j,k})^{2/r-1}}{8},\sqrt{n\left[(r+1)\cH_{j+1}+\lambda+2\ln(j+1)\right]}\right)$, leading to:
\begin{equation*}
    \eps_j t_j\le a_r\eps_j^{2/r}\frac{(B\sigma_1)^{1-\frac{2}{r}}}{n}\left[\cH_{j+1}+\lambda+2\ln(j+1)\right] + b_r\eps_j\sqrt{\frac{\cH_{j+1}+\lambda+2\ln(j+1)}{n}}.
\end{equation*}
The first term is treated in the non-centered case:
\begin{equation*}
    a_r \eps_j^{2/r}\frac{(B\sigma_1)^{1-\frac{2}{r}}}{n}\left[\cH_{j+1}+\lambda+2\ln(j+1)\right] \le A_r\frac{1}{n}\frac{ \deff'\ln(d)+\lambda}{\left(\sigma_1^{-r}B\right)^{1-2/r}}.
\end{equation*}
The second one follows the same lines:
\begin{align*}
    b_r\eps_j\sqrt{\frac{\cH_{j+1}+\lambda+2\ln(j+1)}{n}} & \le b_r'\eps_j\frac{1}{\sqrt{n}} \left(\sqrt{\cH_{j+1}}+\sqrt{\lambda}\right).
\end{align*}
In the same way that we proved $\sum_j \eps_j \cH_j \le C^{te}_r\deff(r)\ln(d)$, we have that, for $C>0$ a constant:
\begin{equation*}
    \sum_{j\ge0}\eps_j \sqrt{\cH_j} \le C \sqrt{\deff(1)\ln(d)},
\end{equation*}
giving us:
\begin{equation*}
     \sum_{j\ge 0}b_r\eps_j\sqrt{\frac{\cH_{j+1}+\lambda+2\ln(j+1)}{n}} \le B_r \frac{\sqrt{\lambda}+\sqrt{\deff(1)\ln(d)}}{\sqrt{n}}.
\end{equation*}
We then have:
$$
\dP\left(Y'\ge \sigma^r\sum_{j\ge 0}\epsilon_j t_{j}\right)\le r\frac{\pi^2}{6} e^{-\lambda},
$$
for:
$$
\sum_{j\ge 0}\epsilon_j t_{j}\le A_r\frac{1}{n}\frac{ \deff'\ln(d)+\lambda}{\left(\sigma_1^{-r}B\right)^{2/r-1}} + B_r \frac{\sqrt{\lambda}+\sqrt{\deff(1)\ln(d)}}{\sqrt{n}}.
$$
for some suitable constant $A_r$ dependent only on $r$. We hence end up with the same computation as in the non-centered case (up to constants), leading to the following result.
For suitable constant $C'_r$, for all $\lambda>0$, one has that:
\begin{equation}
\dP\left(Y'\ge C_r'\sigma_1^r \left(\frac{1}{n}\frac{\lambda+ \deff'\ln(d)}{\left(\sigma_1^{-r}B\right)^{2/r-1}} +  \frac{\sqrt{\lambda}+\sqrt{\deff(1)\ln(d)}}{\sqrt{n}}\right)\right)\le r\frac{\pi^2}{6} e^{-\lambda},
\end{equation}
$Y'$ defined in \eqref{eq:new_bound_2}.

\subsection{Proof of Remark \ref{rq:non_centered_thm}}

\begin{lem}[Maximum of $n$ \emph{i.i.d.}~Subagaussian Random Variables] Let $a_1,...,a_n$ be \emph{i.i.d.}~$\Sigma$-subgaussian random variables. Denote $R=\max_{i=1,...,n}\NRM{a_i}$. There exists a (universal) constant $C>0$ such that, with probability $1-\delta$:
\begin{equation}
    R^2\le 4\sigma_1^2\left(\ln(\delta^{-1})+2\deff(1)+\ln(n)\right).
\end{equation}
\end{lem}
\begin{proof}
Let $t\ge 0$. Using a classical Markov-Chernoff approach, for some $\lambda>0$:
\begin{align*}
    \P(R\ge t) &\le n\P(\NRM{a_1}^2\ge t^2)\\
    &\le ne^{-\lambda t^2/2}\E e^{\lambda \NRM{a_1}^2/2}.
\end{align*}
Then, writing $a_1=\sum_{j=1}^d \mathcal{e}_j\sigma_j X_j$ where $(\mathcal{e_j})_j$ is an orthonormal basis of eigenvectors of $\Sigma$, and $X_1,...,X_d$ are \emph{i.i.d.}~standard gaussian variables $\cN(0,1)$, yields, using independence:
\begin{align*}
    \E e^{\lambda \NRM{a_1}^2/2}&=\prod_{j=1}^d\E e^{\lambda \sigma_j^2 X_j^2/2}\\
    &=\prod_{j=1}^d \frac{1}{\sqrt{1-\lambda\sigma_j^2}},
\end{align*}
where we assume that $\lambda<\sigma_1^{-2}$. We take $\lambda=\frac{1}{2\sigma_1^2}$. Now, using that $\frac{1}{\sqrt{1-u}}\le e^{2u}$ for $0\le u \le 1/2$ yields for this particular $\lambda$:
\begin{align*}
     \E e^{\lambda \NRM{a_1}^2/2}&\le e^{2\sum_{j=1}^d \frac{\sigma_j^2}{\sigma_1^2}}\\
     &= e^{2\deff(1)}.
\end{align*}
We thus get $\P(R\ge t)\le n \exp(-\frac{t^2}{4\sigma_1^2}+2\deff(1))$. For $\delta\in(0,1)$, we hence have that, with probability $1-\delta$:
\begin{equation}
    R^2\le 4\sigma_1^2(\ln(\delta^{-1})+2\deff(1)+\ln(n)).
\end{equation}
\end{proof}

\subsection{Eventual Tightness\label{app:tightness}} 

\subsubsection{Without Centering}
We want to derive possible tightness for our probability bounds. Let $A_1,...,A_n$ \emph{i.i.d.}~centered gaussians of covariance $\Sigma$. Then, for $x_1,...,x_r\in \cS$:
\begin{align*}
    \E\left[\prod_{k=1}^r A_1^\top x_k\right] \le O(\sigma_1^r).
\end{align*}
We then take $x_1=...=x_r=A_1/\NRM{A_1}$ in order to have:
\begin{align}
    \frac{1}{n}\sum_i\prod_{k=1}^r A_i^\top x_k &=\frac{\NRM{A_1}^r}{n} + O(\sigma_1^r)\\
    &= O\left(\frac{\sigma_1^r\deff(1)^{r/2}}{n} +\sigma_1^r\right). \label{eq:tightness_non_centered}
\end{align}
Cotrast this we the results obtained in Theorem \ref{thm:chaining_not_centered}: we require $n$ to be of order $\deff(r)\ln(d)\deff(1)^{r/2-1}$ (Remark \ref{rq:non_centered_thm}) for our bound to be of order $O(1)$. Considerations just above, and in particular \eqref{eq:tightness_non_centered} require $\deff(1)^{r/2}=O(n)$. Our lower and upper bounds match only up to a factor $\frac{\deff(r)\ln(d)}{\deff(1)}$, that should not be too large. However, our dependency in $n$ seems optimal ($1/n$). We believe that $\deff(r)\ln(d)$ instead of $\deff(1)$ is is simply an artifact of the proof.

\subsubsection{With Centering}

We now consider the centered case. The non-centered case suggest that we are not tight in terms of dimension-dependency, we thus restrict ourselves to the dependency in $n$. Consider the same random variables $A_1,...,A_n$ as above. Let, for $x_1,...,x_k\in \cB$:
\begin{equation}
    X=\frac{1}{n}\sum_{i=1}^n \left( \prod_{k=1}^r A_i^\top x_k -\E\left[ \prod_{k=1}^r A_i^\top x_k\right]\right).
\end{equation}
We have:
\begin{align*}
    \E[X^2]=\frac{1}{n}\E\left[ \left( \prod_{k=1}^r A_1^\top x_k -\E \prod_{k=1}^r A_1^\top x_k \right)^2\right].
\end{align*}
We thus observe a dependency in $1/n$ on the second moment. That leads to an optimal dependency in $n$ in our centered bound. Indeed, we have a $1/\sqrt{n}$, but we cannot gain any order of magnitude: if we have something of the form $\P(X\ge \gamma \frac{\lambda+\beta}{n^\alpha})\le e^{-\lambda}$ for all $\lambda>0$, we get $E[X^2]\le \frac{\gamma^2\beta^2}{n^{2\alpha}}$, leading to an optimal exponent $\alpha$ of $1/2$, which we have.

\section{Bounding Random Tensors \label{C-Tensors}\label{app:tensors}: Proof of Theorem \ref{prop:tensors_non_iso_main}}
Given $a_1,...,a_n$ \emph{i.i.d.}~random vectors in $\R^d$, $\Sigma$-subgaussian, we prove that, with probability $1-\delta$:
\begin{equation}
    \NRM{\frac{1}{n}\sum_{i=1}^na_i^{\otimes p} -\esp{a_1^{\otimes p}}}_{\op} \leq \Tilde{\mathcal{O}}\left(\sqrt{\frac{\Tr\big(\Sigma\big)^{2p}}{n}}\right),
\end{equation}
where $\Tilde{\mathcal{O}}$ hides logarithmic factors in $d,\delta^{-1},n$. 
\\

First, recall that:
\begin{equation}
    \NRM{\frac{1}{n}\sum_{i=1}^na_i^{\otimes p} -\esp{a_1^{\otimes p}}}_{\op}=\sup_{x\in\cB}\left\{\frac{1}{n}\sum_{i=1}^n \langle a_i,x\rangle - \esp{\langle a_1,x\rangle }\right\}.
\end{equation}
We begin with a first lemma, that bounds whp the right handside of the above equality, for $x$ fixed.
\begin{lem}\label{lem:moment_p} Let $X_1,...,X_n$ be \emph{i.i.d.} real random variables with 1-subgaussian laws and $p\geq2$. Let $S_n=\sum_{i=1}^n X_i^p$. There exists $C_p>0$ such that for any $\lambda>0$:
\begin{equation}
    \P\left(|S_n-\esp{S_n}|\geq C_p \lambda^{\frac{p+1}{2}}\sqrt{n\ln(n)^p}\right)\leq e^{-\lambda}.
\end{equation}
\end{lem}
\begin{proof}
For $i=1,...,n$, let $Y_i=X_i^p-\esp{X_i^p}$ and for some fixed $R>0$, $Z_i=Y_i \mathds{1}_{|Y_i|<R}$. Denote $S_n^R= \sum_{i=1}^n Z_i$. Since we cannot use Chernoff bounds on the random variables $Y_i$, we artificially bound them by $R>0$, and fix $R$ afterwards. First, using Hoeffding's inequality:
\begin{equation}
    \P(|S_n^R-\esp{S_n^R}|\geq t)\leq 2e^{-\frac{2t^2}{nR^2}}.
\end{equation}
Secondly, if $R>\esp{X_1^p}$,
\begin{align*}
    \P(|Y_1|>R)&\leq \P(X_1^p>R-\esp{X_1^p}) + \P(X_1^p<-R+\esp{X_1^p})\\
    &\leq 2 \P(X_1^p>R-\esp{X_1^p})\\
    &\leq 2 \P(X_1>(R-\esp{X_1^p})^{1/p})\\
    &\leq 2 \exp\left(-\frac{(R-\esp{X_1^p})^{2/p}}{2}\right),
\end{align*}
using the subgaussian tail of $X_1$. This leads to, using a union bound:
\begin{equation}
    \P(\sup_i |Y_i| > R)\leq 2 n\exp\left(-\frac{(R-\esp{X_1^p})^{2/p}}{2}\right).
\end{equation}
Thirdly, 
\begin{align*}
    \esp{S_n}-\esp{S_n^R}&=\esp{S_n\mathds{1}_{\{\forall i, |Y_i|<R}\}^C}\\
    &=n\esp{Y_1\mathds{1}_{|Y_1|>R}}\\
    &\leq n \sqrt{ \P(|Y_1|>R) \esp{Y_1^2}}\\
    &\leq C_p n  \exp\left(-\frac{(R-\esp{X_1^p})^{2/p}}{4}\right).
\end{align*}
We can now prove the desired result. Writing
\begin{align*}
    \P(|S_n-\esp{S_n}|>t)&\leq \P(|S_n-\esp{S_n}|>t,\forall i, |Y_i|<R) + \P(\sup_i |Y_i| > R),
\end{align*}
and noticing that $\P(|S_n-\esp{S_n}|>t,\forall i, |Y_i|<R)=\P(|S_n^R-\esp{S_n}|>t,\forall i, |Y_i|<R)$, making the difference $\esp{S_n}-\esp{S_n^R}$ appear, we have:
\begin{align*}
    \P\big(|S_n-\esp{S_n}|>t\big)&\leq \P\big(|S_n^R-\esp{S_n^R}|>t - (\esp{S_n}-\esp{S_n^R})\big) + \P\big(\sup_i |Y_i| > R\big)\\
    &\leq 2\exp\left( \frac{-\left(t - (\esp{S_n}-\esp{S_n^R})\right)^2}{nR^2}\right) + 2 n\exp\left(-\frac{(R-\esp{X_1^p})^{2/p}}{2}\right).
\end{align*}
We choose $R>2\esp{X_1^p}$ such that $2n  \exp\left(-\frac{(R-\esp{X_1^p})^{2/p}}{4}\right)=e^{-u}$ for some $u\geq0$. Then, $\esp{S_n}-\esp{S_n^R}\leq c_p$ and:
\begin{align*}
    \P\big(|S_n-\esp{S_n}|>t\big)&\leq C_p e^{-\frac{t^2}{nR^2}} +e^{-u},
\end{align*}
where $u=\frac{(R-\esp{X_1})^{2/p}}{4}-\log(2n)$. Now, setting $\frac{t^2}{nR^2}=u$ yields:
\begin{align*}
    t^2&=unR^2\\
    &=C_p'nu(u+\log(n))^p,
\end{align*}
and thus:
\begin{equation}
    \P\big(|S_n-\esp{S_n}|> C_p'' \sqrt{nu(u+\log(n))^p} \big)\leq C_p''' e^{-u}.
\end{equation}
The condition $R>2\esp{X_1^p}$ is translated in $u>c_p$ that can be taken into account in constant $C_p'''$.
\end{proof}

Using this lemma, we can now prove the following:
\begin{thm} Let $T=\frac{1}{n}\sum_{i=1}^na_i^{\otimes p}$ for $p\geq 2$, where $a_1,...,a_n$ are $\Sigma$-subgaussian random variables in $\R^d$. There exists a constant $C_p>0$ such that for any $\delta\in(0,1)$, with probability at least $1-\delta$, we have:
\begin{equation}
        \NRM{T-\E T}_{\op} \leq C_{p}\sigma_1^p\sqrt{\frac{\big(\deff+\ln(d)+\ln(\delta^{-1})\big)^{p+1}\ln(n)^p}{n}}.
\end{equation}
where $\sigma_1^2$ is the largest eigenvalue of $\Sigma$ and $\deff=\frac{\Tr(\Sigma)}{\sigma_1^2}$.
\end{thm}

\begin{proof}
Remind that we have:
\begin{align*}
    \NRM{\frac{1}{n}\sum_{i=1}^na_i^{\otimes p} -\esp{a_1^{\otimes p}}}_{\op} =\sup_{x\in\cB}\left\{\frac{1}{n}\sum_{i=1}^n \langle a_i,x\rangle - \esp{\langle a_1,x\rangle }\right\},
\end{align*}
and thus, writing $T=\frac{1}{n}\sum_{i=1}^na_i^{\otimes p}$:
\begin{align*}
    \NRM{T-\esp{T}}_\op &= \frac{1}{n} \sup_{x \in\cS} \left\{\sum_{i=1}^n\langle \sqrt{\Sigma}^{-1}a_1,\sqrt{\Sigma}x\rangle^p-\esp{\langle \sqrt{\Sigma}^{-1}a_i,\sqrt{\Sigma}x\rangle^p}\right\}\\
    &= \frac{1}{n} \sup_{\sqrt{\Sigma}^{-1}y \in\cS} \left\{\sum_{i=1}^n\langle \sqrt{\Sigma}^{-1}a_i,y\rangle^p-\esp{\langle \sqrt{\Sigma}^{-1}a_i,y\rangle^p}\right\},
\end{align*}
and the $(\sqrt{\Sigma^{-1}}a_i)_{i=1,...,n }$ are \emph{i.i.d.} distributed and $I_d$-subgaussian. Denote $Z_y= \sum_{i=1}^n\langle \sqrt{\Sigma}^{-1}a_i,y\rangle^p$, for any $y\in\sqrt{\Sigma}\cS$. We know that for such $y$, we have $\NRM{y}\le \sigma_1$. Using Lemma \ref{lem:moment_p} for $y\in\sqrt{\Sigma}\cS$:
\begin{equation*}
    \P\left(\frac{1}{n}(Z_y-\E Z_y) \ge C_p\sigma_1^p\sqrt{\frac{\lambda^{p+1}\ln(n)^p}{n}}\right)\le e^{-\lambda}.
\end{equation*}
Let now $\cN$ be an $1/(2p\sigma_1^p)$-covering of $\sqrt{\Sigma}\cS$. We thus have:
\begin{equation*}
        \P\left(\sup_{x\in\cN} \langle(T-\E T),x^{\otimes p}\rangle \ge C_p\sigma_1^p\sqrt{\frac{\lambda^{p+1}\ln(n)^p}{n}}\right)\le e^{-\lambda + \ln(|\cN|)}.
\end{equation*}

\paragraph{Isotropic case ($\Sigma=I_d$)} In the isotropic case, one can achieve $\ln(|\cN|)\leq d\ln(1+4p)$ \citep{rogers_1963}. In that case, with probability $1-\delta$ we have:
\begin{equation}
    \sup_{x\in\cN} \langle(T-\E T),x^{\otimes p}\rangle\leq C_p\sigma_1^p\sqrt{\frac{\big(d+\ln(\delta^{-1})\big)^{p+1}\ln(n)^p}{n}}.
\end{equation}
Now, let $y\in\cS$. There exists $x\in\cN$ such that $\NRM{x-y}\le 1/(2p)$. 
\begin{align*}
    \langle(T-\E T),y^{\otimes p}\rangle &\le \langle(T-\E T),x^{\otimes p}\rangle + \NRM{T- \E T}_\op \NRM{x^{\otimes p}-y^{\otimes p}}\\
    &\le \sup_{x\in\cN} \langle(T-\E T),x^{\otimes p}\rangle + \frac{1}{2}\NRM{T-\E T}_\op,
\end{align*}
using the fact that for $x,y\in\cS$ we have $\NRM{x^{\otimes p}-y^{\otimes p}}\leq p\NRM{x-y}\leq 1/2$ (Lemma 10 in \cite{bubeck2020law}). This concludes the proof in the isotropic case, by taking a supremum over $y\in\cS$, leading to:
\begin{equation*}
    \langle(T-\E T),y^{\otimes p}\rangle \le 2\sup_{x\in\cN} \langle(T-\E T),x^{\otimes p}\rangle.
\end{equation*}

\paragraph{Non-isotropic case (general $\Sigma$)}
Let us use results from \cite{pmlr-v134-even21a} (Section 3) in order to bound $\ln(|\cN|)$. For any $r\geq1$,
\begin{equation*}
    \ln(|\cN|)\le C_{p,r}'\left[ \min\left(d-1,(2p)^{2/r}(\deff(r)-1)/e\right) \ln\left(\max\left(e,(2p)^{2/r}\frac{\deff(r)-1}{d-1} \right)\right)+\ln(d)+\deff(r)\right],
\end{equation*}
and thus, if we assume that $(2p)^{2/r}\le e\frac{d-1}{\deff(r)-1}$, we have:
\begin{equation*}
    \ln(|\cN|)\le C_{p,r}' \left((2p)^{2/r}\frac{\deff(r)}{e}+\ln(d)\right).
\end{equation*}
Back to our probabilistic bound:
\begin{equation*}
        \P\left(\sup_{x\in\cN} \langle(T-\E T),x^{\otimes p}\rangle \ge C_p\sigma_1^p\sqrt{\frac{\lambda^{p+1}\ln(n)^p}{n}}\right)\le e^{-\lambda + C_{p,r}' \left((2p)^{2/r}\frac{\deff(r)}{e}+\ln(d)\right)}.
\end{equation*}
yielding the stated result, using the same argument as in the isotropic case: for some constant $C_{p,r}$, for any $\delta>0$, with probability $1-\delta$,
\begin{equation*}
    \NRM{T-\E T}_{\op} \leq C_{p,r}\sigma_1^p\sqrt{\frac{\big(\deff(r)+\ln(d)+\ln(\delta^{-1})\big)^{p+1}\ln(n)^p}{n}}.
\end{equation*}
Let now take $r=1$. If $(2p)^{2}\le e\frac{d-1}{\deff(1)-1}$, we have with probability $1-\delta$:
\begin{equation*}
    \NRM{T-\E T}_{\op} \leq C_{p,1}\sigma_1^p\sqrt{\frac{\big(\deff(1)+\ln(d)+\ln(\delta^{-1})\big)^{p+1}\ln(n)^p}{n}}.
\end{equation*}
If $(2p)^{2}> e\frac{d-1}{\deff(1)-1}$, we have:
\begin{equation*}
    \NRM{T-\E T}_{\op} \leq C_{p,\infty}\sigma_1^p\sqrt{\frac{\big(\deff(1)+\ln(d)+\ln(\delta^{-1})\big)^{p+1}\ln(n)^p}{n}}.
\end{equation*}
In the latter case, since $d\leq c_p\deff(1)$, we still have
\begin{equation*}
    \NRM{T-\E T}_{\op} \leq C_{p}' \sigma_1^p\sqrt{\frac{\big(\deff(1)+\ln(d)+\ln(\delta^{-1})\big)^{p+1}\ln(n)^p}{n}}.
\end{equation*}
Finally, with $C_p=\max(C_p',C_{p,1})$, in both cases, with probability $1-\delta$
\begin{equation*}
    \NRM{T-\E T}_{\op} \leq C_{p}\sigma_1^p\sqrt{\frac{\big(\deff(1)+\ln(d)+\ln(\delta^{-1})\big)^{p+1}\ln(n)^p}{n}}.
\end{equation*}

\end{proof}

\section{Statistical Preconditioning}
\subsection{Large Deviation of Hessians  and Proposition \ref{prop:deviation_hess}}
\begin{proof} We first apply Theorem \ref{thm:chaining_centered} and Remark \ref{rq:non_centered_thm} with $f_1=f_2=Id$ and $f_3=\ell'' -\ell''(0)$, giving a bound on the following quantity:
\begin{equation*}
    M:=\sup_{x,y,z\in\cB} \frac{1}{n}\sum_{i=1}^n\left\{ (a_i^\top x)( a_i^\top y )(\ell''(a_i^\top z) - \ell''(0))- \E\left[(a_i^\top x)( a_i^\top y) (\ell''(a_i^\top z) - \ell''(0))\right]\right\}.
\end{equation*}
Now, notice that:
\begin{equation*}
    \NRM{H_x-\Bar{H}_x}_\op \le M + N,
\end{equation*}
where 
\begin{equation*}
    M':=\sup_{x,y\in\cB}\frac{\ell''(0)}{n} \sum_{i=1}^n\left\{ (a_i^\top x)( a_i^\top y) - \E\left[(a_i^\top x)( a_i^\top y) \right]\right\}.
\end{equation*}
Again, $M'$ can be bounded by Theorem \ref{thm:chaining_centered} and Remark \ref{rq:non_centered_thm}.
\end{proof}

\subsection{Bregman Gradient Descent: Algorithms and Theoretical Guarantees}

\noindent \textbf{Problem Formulation:} As mentioned in Section \ref{section:statistical_precond_general}, we aim at solving the following problem:
\begin{equation}
    \min_{x\in\R^d} \Phi(x):= F(x)+\psi(x), 
\end{equation}
for some convex regularizer $\psi$ on a convex domain $\Dom_\psi$, and $F$ $\sigma_{F/\phi}$ relatively strongly convex and $L_{F/\phi}$ relatively smooth with respect to some strongly convex function $\phi$ (named the preconditioner). We still denote $\kappa_{F/\phi}=\frac{\sigma_{F/\phi}}{L_{F/\phi}}$ their relative condition numbers.
\\

\noindent \textbf{Bregman Gradient Descent:} The most classical algorithm in order to solve this optimization problem is \emph{Bregman Gradient Descent} or \emph{Mirror Gradient Descent}. The algorithm is the following, as sketched in Section \ref{section:statistical_precond_general}.
\begin{enumerate}
    \item Start from $x_0\in\Dom_\psi$;
    \item For $t\in\N$ and some stepsize $\eta_t>0$, perform the update:
    \begin{equation}
    x_{t+1}\in \argmin_{x\in \R^d} \left \{ \langle \nabla F(x_t),x \rangle + \psi(x) +\frac{1}{\eta_t} D_\phi(x,x_t) \right \}.\label{eq:bregman_gradient_simple_app}
\end{equation}
\end{enumerate}
For $\phi=\frac{\NRM{.}^2}{2}$, we get classical proximal gradient descent.
\begin{prop}[Bregman Gradient Descent: Convergence Guarantees] For stepsizes $\eta_t=\frac{1}{L_{F/\phi}}$ and $x^*$ the minimizer, \eqref{eq:bregman_gradient_simple_app} yields:
\begin{equation*}
    D_\phi(x_t,x^*)\le \left(1-\kappa_{F/\phi}^{-1}\right)^t D_\phi(x_0,x^*).
\end{equation*}
If $\phi$ is $\mu$-strongly convex, one has:
\begin{equation*}
    \NRM{x_t-x^*}^2\le\frac{1}{\mu}\left(1-\kappa_{F/\phi}^{-1}\right)^t D_\phi(x_0,x^*).
\end{equation*}
\end{prop}
\begin{proof} For simplicity, we only assume that $\psi=0$ (no regularization). 
Let $V_t(x)=\langle \nabla F(x_t),x \rangle +\frac{1}{\eta_t} D_\phi(x,x_t)$. One has $\nabla V_t(x)=\nabla F(x_t)+\frac{1}{\eta_t}(\nabla \phi(x)-\nabla\phi(x_t)).$ As $\nabla V_t(x_{t+1})=0$, we have:
\begin{equation*}
    \eta_t\nabla F(x_t)+ \nabla \phi (x_{t+1})-\nabla \phi(x_t) =0.
\end{equation*}
Moreover:
\begin{equation*}
    V_t(x^*)-V_t(x_{t+1})=D_\phi(x_{t+1},x^*),
\end{equation*}
leading to:
\begin{equation*}
    D_\phi(x_{t+1},x^*)=\eta_t\nabla F(x_t)^\top(x^*-x_{t+1})+D_\phi(x_t,x^*)-D_\phi(x_{t+1},x_t).
\end{equation*}
In order to study $\eta_t\nabla F(x_t)^\top(x^*-x_{t+1})$, we write:
\begin{equation*}
    \eta_t\nabla F(x_t)^\top(x^*-x_{t+1})=\eta_t\nabla F(x_t)^\top(x^*-x_{t})+\eta_t\nabla F(x_t)^\top(x_t-x_{t+1}).
\end{equation*}
We have $\eta_t\nabla F(x_t)^\top(x^*-x_{t})=\eta_t(F(x^*)-F(x_t)-D_F(x_t,x^*))$. For the second term, we remark that;
\begin{align*}
    D_\phi(x_{t+1},x_t)+D_\phi(x_t,x_{t+1})&=\langle \nabla\phi(x_t)-\nabla\phi(x_{t+1}),x_t-x_{t+1}\rangle\\
    &=\eta_t\langle \nabla F(x_t),x_t-x_{t+1}\rangle.
\end{align*}
Plugging all this leads to:
\begin{align*}
    D_\phi(x_{t+1},x^*)&=D_\phi(x_t,x^*)-\eta_t(D_F(x^*,x_t)+F(x_t)-F(x^*))+D_\phi(x_t,x_{t+1})\\
    &\le (1-\eta_t\sigma_{F/\phi})D_\phi(x^*,x_t)+D_\phi(x_t,x_{t+1})-\eta_tD_F(x_t,x^*).
\end{align*}
Finally, using Bregman co-coercivity yields $D_\phi(x_t,x_{t+1})\le\eta_tD_F(x_t,x^*)$ for $\eta_t=1/L_{F/\phi}$, hence the result.
\end{proof}

\noindent \textbf{Acceleration and \emph{SPAG} Algorithm:} although \cite{dragomir2019optimal} prove that the rate of convergence $(1-\kappa_{F/\phi})$ above is optimal, \cite{hendrikx2020statistically} propose an acceleration (\emph{SPAG} algorithm) in the sense that asymptotically, one can reach a rate of convergence $(1-\sqrt{\kappa_{F/\phi}})$. Using stochastic gradients is also possible \citep{dragomir2021fast}, with or without variance reduction.

\subsection{Bounding Condition Numbers and Consequences on Statistical Preconditioning}

For $\mu>0$ such that $ \forall x \in \Dom_\psi, \NRM{\nabla^2 f (x)-\nabla^2F(x)}_\op \le \mu$, we have, for all $x\in\Dom_\psi$ and for $\phi(x)=f(x)+\frac{\NRM{x}^2}{2}$ (inequalities are taken in terms of symmetric matrices):
\begin{align*}
    \nabla^2 F(x)\le \nabla^2 f(x) + \mu I_d=\nabla^2 \phi(x),
\end{align*}
giving us $L_{F/\phi}\le 1$. Then, for relative strong*convexity, as $F$ is $\lambda$-strongly convex:
\begin{align*}
    \nabla^2f(x)+\mu I_d &\le \nabla^2 F(x)+2\mu I_d\\
    &\le (1+2\mu/\lambda) \nabla^2 F(x).
\end{align*}
Hence, we obtain:
\begin{equation}
    \kappa_{F/\phi}\le 1+\frac{2\mu}{\lambda}.
\end{equation}
Proposition \ref{prop:mu_precond} bounds this $\mu$ with high probability using large deviations on Hessians, in order to apply these considerations.

\subsection{Assumption \ref{hyp:erm} Encompasses Logistic and Ridge Regressions \label{app:ERM}}

The statistician has access to feature vectors $a_1,...,a_n$, and corresponding labels $b_1,...,b_n$. Linear models (including logistic and ridge regression) take the form $\ell(x, (a_i,b_i))=\ell_i(a_i^\top x)+\frac{\lambda}{2}\NRM{x}^2$. Linear regression problems then reduces to the minimization of:
\begin{equation}
    \frac{1}{n} \sum_{i=1}^n \ell_i(a_i^\top x) +\frac{\lambda}{2}\NRM{x}^2.
\end{equation}
It is then to be noticed that for logistic and ridge regressions, functions $\ell_i$ verify $\ell_i''=\ell_j''$ for $i,j\in [n]$.

\section{Randomized Smoothing\label{app:smooth}}
\subsection{Randomized Smoothing: Detailed Algorithm and Convergence Guarantees for General Smoothing Distributions}

\noindent \textbf{Detailed Algorithm:} we here describe in details how the algorithm works. We recall that $f(x)=\E_a[F(x,a)]$ for $x\in \R^d$. $\mu$ is the smoothing distribution, $\phi$ the known regularizing function. The algorithm uses three sequences of points $(x_t,y_t,z_t)_t$, where $y_t$ is the \emph{query} point: at iteration $t$, stochastic gradients are computed using $y_t$.  The three sequences evolve according to a dual-averaging algorithm, involving three scalars $L_t,\theta_t,\eta_t$ to control the stepsizes. The smoothed gradients use a sequence of scalars $(u_t)_t$. The algorithm:
\begin{enumerate}
    \item Computes $y_{t} = (1-\theta_t)x_t + \theta_t z_t$.
    \item Draws $Z_{1,t},...,Z_{m,t}$ \emph{i.i.d.}~random variables according to the smoothing distribution $\mu$, for $m$ a fixed integer.
    \item Queries the oracle at the m points $y_t+u_t Z_{i,t},i=1,...,m$, yielding stochastic gradients $g_{i,t}\in \partial F(y_t+u_t Z_{i,t},a_{i,t})$.
    \item Computes the average $g_t=\frac{1}{m} \sum_{i=1}^m g_{i,t}$.
    \item Performs the update:
    \begin{equation}
        \begin{array}{ccc}
            z_{t+1} &=& \argmin_x \left\{ \sum_{\tau=0}^t \frac{1}{\theta_\tau}\langle g_\tau,x\rangle +\sum_{\tau=1}^t \frac{1}{\theta_\tau} \phi(x) +\frac{1}{2}(L_{t+1} + \frac{\eta_{t+1}}{\theta_{t+1}})\NRM{x}^2\right\},  \\
            x_{t+1} &=& (1-\theta_t)x_t+\theta_t z_{t+1}.
        \end{array} 
    \end{equation}
\end{enumerate}
\noindent \cite{duchi2012randomized} obtain the folowing result.
\begin{prop}[Convergence Guarantees for General Smoothing]\label{prop:general_smooth} Assume that there exist constants $L_0$ and $L_1$ such that for all $u>0$, we have $\E_{Z\sim\mu}[f(x+uZ)]\le f(x)+L_0u$, and $\E_{Z\sim\mu}[f(x+uZ)]$ has $L_1$-Lispchitz continuous gradient. Set $u_t=\theta_t u$, $L_t=L_1/u_t$, and assume that $\eta_t$ is non-decreasing. Set $\theta_0=1$, and $\theta_{t+1}=\frac{2}{1+\sqrt{1+4/\theta_{t-1}^2}}$. Assume that $\NRM{x^*}\le R$. Then, for all $T>0$:
\begin{equation}
    \E[f(x_T)+\phi(x_T)-f(x^*)-\phi(x^*)]\le \frac{6L_1R^2}{Tu}+\frac{2\eta_TR^2}{T}+\frac{1}{T}\sum_{t=0}^{T-1}\frac{1}{\eta_t}\E[\NRM{e_t}^2]+\frac{4L_0u}{T},
\end{equation}
where $e_t=\nabla f_{\mu_t}(y_t)-g_t$ is the error in the gradient estimate.
\end{prop}

\subsection{Isotropic Smoothing: Proof of both Propositions \ref{prop:iso_smooth} and \ref{prop:iso_smoothing_conv}} 

In the isotropic case, $\mu=\cN(0,I_d)$ is the smoothing distribution. We now assume that $F(.,a)$ and thus $f$ are $L$-Lipschitz. Proposition \ref{prop:iso_smooth} leads to explicit constants $L_0$ and $L_1$ in Proposition \ref{prop:general_smooth} just above. We prove Proposition \ref{prop:iso_smooth} here.\\
\begin{proof}
For all $x\in\R^d$, one has with Jensen inequality:
\begin{equation*}
    f(x)\le f^\gamma(x).
\end{equation*}
Then, we obtain $f^\gamma(x)\le f(x)+\gamma L \sqrt{d}$ using Lipschitz continuity of $f$ and $\E\NRM{Z}\le \sqrt{d}$. In order to prove that $f^\gamma$ is $L/\gamma$-smooth, we need to compute its gradient:
\begin{align*}
    \frac{f^\gamma(x+h)-f^\gamma(x)}{h}&= \int_{\R^d}{\rm d}z f(z)(\mu(z-h/\gamma)-\mu(z))/h\\
    &= -\frac{1}{\gamma} \E_Z[f(x+\gamma Z)Z] 
\end{align*}
when $h\to 0$, where $\mu(z)$ is the density of the smoothing distribution. As in \cite{duchi2012randomized}, we then have that:
\begin{equation*}
    \NRM{\nabla f^\gamma(x)-\nabla f^\gamma (y)} \le \frac{1}{\gamma} \int_{\R^d}{\rm d}z L_0|\mu(z-x)-\mu(z-y)|.
\end{equation*}
The end of the proof follows as in their Lemma 10.
\end{proof}

\noindent Using these properties, and setting $\eta_t=\frac{L\sqrt{t+1}}{R\sqrt{m}}$, $u=Rd^{-1/4}$ and $L_t=L/u_t$, Proposition \ref{prop:iso_smoothing_conv} is obtained by simplifying the expression in Proposition \ref{prop:general_smooth} \citep{duchi2012randomized}.

\subsection{Non-Isotropic Smoothing}

We now focus on non-isotropic smoothing distributions: $\mu=\cN(0,\Sigma')$, for $\Sigma'$ a symmetric definite positive matrix to determine. We start by proving Proposition \ref{prop:smooth_prop_non_iso}.
\begin{proof}
Let $X\sim \mathcal{N}(0,\Sigma')$. Denote, for $i\in [N]$:
\begin{equation}
    f_i^\gamma(x)=\E[\ell(a_i^\top(x+\gamma X))].
\end{equation}
We have using $L$-Lipschitz continuity of $f$:
\begin{equation}
    f\le f^\gamma \le f+\gamma L \sqrt{\deff(1)\sigma_1^2}\sqrt{\Tr(\Sigma')}.
\end{equation}
Some computations lead to: $f_i^\gamma$ is differentiable and
\begin{align}
    \nabla f_i^\gamma(x)&=-\frac{1}{\gamma}\E[\ell_i(a_i^\top(x+\gamma X))\Sigma'^{-1}X]\\
    &=-\frac{1}{\gamma}\E[\ell_i(a_i^\top(x+\gamma \Sigma'Y))Y] \text{ where }Y\sim\mathcal{N}(0,\Sigma'^{-1})\\
    &=-\frac{1}{\gamma}\E[\ell_i(a_i^\top(x+\gamma \Sigma'Y))a_i \times \frac{\langle Y,a_i\rangle}{\NRM{a_i}^2}],
\end{align}
as only the contribution of $Y$ in $\R a_i$ is to be taken into account. The form \eqref{eq:var_without_centering} begins to appear here.
\begin{align*}
    \NRM{ \nabla f^\gamma(x)- \nabla f^\gamma(y)}&=\frac{1}{\gamma}\sup_{\NRM{v}\le 1} \frac{1}{N}\sum_{i=1}^N \E_Y[(\ell_i(a_i^\top(x+\gamma \Sigma'Y))-\ell_i(a_i^\top(y+\gamma \Sigma'Y)))  \frac{a_i\langle Y,a_i\rangle}{\NRM{a_i}^2}]^\top v\\
    &\le \frac{1}{\gamma}\sup_{\NRM{v}\le 1} \frac{1}{N}\sum_{i=1}^N \E_Y[L|a_i^\top(x-y)| |a_i^\top v| \frac{|\langle Y,a_i\rangle|}{\NRM{a_i}^2}]\\
    &=\frac{1}{\gamma}\sup_{\NRM{v}\le 1} \frac{1}{N}\sum_{i=1}^N L|a_i^\top(x-y)| |a_i^\top v| \frac{\E_Y[|\langle Y,a_i\rangle|]}{\NRM{a_i}^2}\\
    &\le\frac{1}{\gamma}\frac{\max_i \NRM{a_i}_{\Sigma'^{-1}}}{\min_i\NRM{a_i}^2}\sup_{\NRM{v}\le 1} \frac{1}{N}\sum_{i=1}^N L|a_i^\top(x-y)| |a_i^\top v|.
\end{align*}
Using Theorem \ref{prop:tensors_non_iso_main}, we have that:
\begin{equation*}
    \NRM{\frac{1}{N}\sum_ia_ia_i^{\top}-\E a_ia_i^{\top}}_{\op}\leq C\sqrt{\frac{\deff(1)+\ln(\delta^{-1})}{N}}.
\end{equation*}
Thus, with probability $1-\delta$:
\begin{align*}
    \sup_{\NRM{v}\le 1,\NRM{x-y}\le1} \frac{1}{N}\sum_{i=1}^N L|a_i^\top(x-y)| |a_i^\top v| &\leq C\sigma_1^2\sqrt{\frac{\deff(1)+\ln(\delta^{-1})}{N}} + \sigma_1^2.
\end{align*}
Furthermore, tightness of norms of gaussian random variables around their mean lead to $\min_i\NRM{a_i}^2\approx \sigma_1^2 \deff(1)$. Then, $\max_i \NRM{a_i}_{\Sigma'^{-1}}\approx \Tr(\Sigma \Sigma'^{-1})$. Minimizing this under $\NRM{\Sigma'}=C^{te}$ leads to $\Sigma'=\sqrt{\Sigma}$.\\

\noindent All in one, with $\Sigma'=\sqrt{\Sigma}$, we end up with:
\begin{align*}
    f^\gamma & \le f + \gamma L \sqrt{\sigma_1^3\deff(1)\deff(2)},   \\
    \NRM{\nabla f^\gamma}_{\Lip} & \le  \frac{L\sigma_1^{1/2}\deff(2)^{1/2}}{\gamma\deff(1)}\left(1+C\sqrt{\frac{\deff(1)\ln(d)+\ln(\delta^{-1})}{N}}\right).
\end{align*}
\end{proof}
Then, Proposition \ref{prop:non_iso_smooth_conv} is obtained in the same way as Proposition \ref{prop:iso_smoothing_conv}, setting $\eta_t=\frac{L\sqrt{t+1}}{R\sqrt{m}}$, $u=\frac{R}{\frac{L\sigma_1^{1/2}\deff(2)^{1/2}}{\gamma\deff(1)}\left(1+C\sqrt{\frac{\deff(1)\ln(d)+\ln(\delta^{-1})}{N}}\right)}$ and $L_t=L/u_t$. We just replaced $d^{-1/4}$ by a less ergonomic, yet smaller expression.

\section{Robustness of Two-Layered Neural Networks with Polynomial Activation \label{F-robustness}}
In this section, we present two applications of our chaining bounds, that played a role of toy problem.
\cite{bubeck2020law} conjecture that two-layered neural networks interpolating \emph{generic data} (defined below)) have a Lipshitz constant that must be lower bounded by $\sqrt{\frac{n}{k}}$ where $n$ is the number of data points, and $k$ the number of neurons.
\begin{defn}[Generic Data 1]
Data $(x_i,y_i)_{1\le i \le n}$ are generic if they are \emph{i.i.d.}~and if $y_i$ are centered random signs, $x_i$ centered gaussians of covariance $I_d/d$.
\end{defn}
\noindent We aim at generalizing some of their results in a non-isotropic framework. We thus define in another way generic data. 
\begin{defn}[Generic Data 2] \label{def:gen_data_2}
Data $(x_i,y_i)_{1\le i \le n}$ are generic if they are \emph{i.i.d.}~and if $y_i$ are centered random signs, $x_i$ centered gaussians of covariance $\Sigma/(\sigma_1^2\deff(1))$.
\end{defn}

\begin{defn}[Tensor] A tensor of order $p\in\N^*$ is an array $T=(T_{i_1,...,i_p})_{i_1,...,i_p\in [d]}\in \R^{dp}$.

\noindent $T$ is said to be of rank $1$ if it can be written as:
$$T=u_1\otimes \cdots \otimes u_p$$ for some $u_1,...,u_p \in \R^p$.

\noindent Scalar product is defined as:
\begin{equation*}
    \langle T, S \rangle = \sum_{i_1,...,i_p} T_{i_1,...,i_p}S_{i_1,...,i_p}.
\end{equation*}
\noindent We define the operator norm of a tensor as:
\begin{equation*}
    \NRM{T}_{\op}=\sup_{\NRM{x_1\otimes...\otimes x_p}\le 1} \langle T, x_1\otimes...\otimes x_p \rangle.
\end{equation*}
\end{defn}
\begin{defn}[Two Layered Neural Network] A two-layered neural network with inputs in $\R^d$, $k$ neurons and Lipschitz non-linearity $\psi$ is a function of the form:
\begin{equation}
    f(x)=\sum_{\ell=1}^k a_\ell \psi(w_\ell^\top x + b_\ell).
\end{equation}
\end{defn}
\noindent \textbf{Conjecture:} A two-layered neural network $f$ that fits generic data $(x_i,y_i)_{1\le i \le n}$ must satisfy, with high probability when $d\to \infty$, for some constant $c>0$ \citep{bubeck2020law}:
\begin{equation}
    \Lip_{\cS}(f)\ge c \sqrt{\frac{n}{k}}.
\end{equation}
That conjecture is not proven (just in some very particular cases and regimes). However, we propose to adapt considerations made with polynomial activation functions $\psi$ in the isotropic regime, to the non-isotropic one. Our aim is however not to link $k$ the number of neurons, to $n$ the number of observations. Indeed, we believe that in this model of \emph{generic data} (both isotropic and non-isotropic ones), dimensionality plays a core role in the Lipschitz constant of $f$. If one considers different dimensions, $n$ and $k$ being fixed, it is natural to believe that, due to the concentration of gaussians in small dimensions, the Lipschitz constant will be bigger for smaller dimensions. Furthermore, adding non-isotropy and introducing effective dimensions should not change this replacing dimensions by effective ones, hence the following proposition, which aims at giving insights on the impact of (effective) dimension on the Lipschitz constant of $f$.
\begin{prop} Assume that $(x_i,y_i)_{1\le i\le n}$ are generic data (Definition \ref{def:gen_data_2}).
Let $\psi(t)=\sum_{q=0}^p\alpha_qt^q$ and $f$ a two-layered neural network with activation function $\psi$ such that $\forall i, f(x_i)=y_i$. Then, if $\deff(1)\le c_0n$ for some $c_0>0$, with probability $1-a\exp(-b\deff(1))$:
\begin{equation} 
\Lip_{\cS}(f)\ge C_p \frac{n}{d^{p-1}\deff(1)}.
\end{equation}
\end{prop}
Either the bound is not tight (likely), or achieving better Lipschitz constants for $f$ is easier with non isotropic data. Both are possible however, and suggest the importance of effective dimensions in the robustness of neural networks. The proof below follows the same steps as in \cite{bubeck2020law}.
\begin{proof} Note that there exist $T_0,...,T_p$ tensors such that $T_q$ is of order $q$ and:
\begin{equation*}
    f(x)=\sum_q \langle T_q , x^{\otimes q}\rangle.
\end{equation*}
Let $\Omega_q=\sum_{i=1}^n y_i x_i^{\otimes q}$. We have:
\begin{equation*}
    n=\sum_i y_i f(x_i)=\sum_{q=0}^p \langle T_q ,\Omega_q\rangle.
\end{equation*}
Hence, there exists $q\ge 1$ such that $\langle T_q ,\Omega_q\rangle\ge c_p n$.
\begin{align*}
    c_p n&\le \langle T_q ,\Omega_q\rangle\\
    &\le \NRM{\Omega_q}_{\op} \NRM{T_q}_{op,*}\\
    &\le d^{q-1} \NRM{\Omega_q}_{\op} \NRM{T_q}_{op},
\end{align*}
using that $ \NRM{T_q}_{op,*}\le d^{q-1}  \NRM{T_q}_{op}$. We then notice that:
\begin{equation*}
     \frac{1}{n}\NRM{\Omega_q}_{\op} = \sup_{z_1,...,z_q\in \cS}\frac{1}{n} \sum_{i=1}^n y_i \prod_{k=1}^q x_i^\top z_k,
\end{equation*}
which is exactly the same form as $Y$ in \eqref{eq:var_without_centering}, except for the $y_i$'s. However, we need a centered bound here: we will use the $y_i$ for this. Let $n^+=\{i:y_i=1\}$, and $n^-=\{i:y_i=-1\}$. We have:
\begin{align*}
    \frac{1}{n} \sum_{i=1}^n y_i \prod_{k=1}^q x_i^\top z_k&=\frac{1}{n} \left(\sum_{i\in n^+}y_i \prod_{k=1}^q x_i^\top z_k -\E\left[\prod_{k=1}^q x_1^\top z_k\right] \right)\\
    &+\frac{1}{n} \left(\sum_{i\in n^-} y_i\prod_{k=1}^q x_i^\top z_k +\E\left[\prod_{k=1}^q x_1^\top z_k\right]\right)\\
    &+\frac{n^+-n^-}{n}\E\left[\prod_{k=1}^q x_1^\top z_k\right].
\end{align*}
With probability $1-C\exp(-c\tau)$ (with respect to the $y_i$'s):
\begin{equation*}
    |n^+-n^-|\le \sqrt{n\tau}.
\end{equation*}
We then have with probability $1-Ce^{-c\tau}$:
\begin{equation*}
    \frac{n^+-n^-}{n}\E\left[\prod_{k=1}^q x_1^\top z_k\right]\leq \sqrt{\frac{\tau}{n}}.
\end{equation*}
The first two terms can be bounded using Theorem \ref{prop:tensors_non_iso_main} and the fact that under the event $|n^+-n^-|\le \sqrt{n\tau}$, we have $n^+,n^-\geq \frac{n-\sqrt{n\tau}}{2}\geq n/3$ if $\tau \le c_0n$.
Using these considerations gives with probability $1-Ce^{-c\tau}-4\delta$:
\begin{align*}
    \NRM{\Omega_q}_{\op}\le \sqrt{\frac{\tau}{n}} +2c'\sqrt{\frac{\ln(\delta^{-1})+\deff(1)}{n}}.
    %\frac{\ln(\delta^{-1})+\deff(q)\ln(d)}{n\deff(1)} + \frac{\sqrt{\ln(\delta^{-1})}+\sqrt{\deff(q)\ln(d)}}{\sqrt{n}\deff(1)^{q/2}}\right).
\end{align*}
Taking $\delta=\exp(-\deff(1))$ and $\tau=\deff(1)$ yields:
\begin{align*}
   \frac{1}{n} \NRM{\Omega_q}_{\op}\le C'\sqrt{\frac{\deff(1)}{n}}.
\end{align*}
Finally, we have, with probability $1-a\exp(-\deff(1))$, if $\deff(1)\le c_0 n$:
\begin{align*}
    \NRM{T_q}_{\op}&\ge \frac{c_p n}{d^{q-1}\NRM{\Omega_q}_{\op}}\ge C_q \frac{n}{d^{q-1}\deff(1)}.
%    & \ge C_q\min\left( \frac{n\deff(1)}{d^{q-1}\deff(q)\ln(d)}, \frac{\sqrt{n\deff(1)^{q/2}}}{d^{q-1}\sqrt{\deff(q)\ln(d)}}\right).
\end{align*}
Then, by observing that the Lipschitz constant of $f$ on the unit ball is lower bounded by $\NRM{T_q}_{\op}$ for any $q$ (with a constant multiplicative factor, using Markov brother's inequality), we obtain with probability $1-a\exp(-b\deff(p)\ln(d))$:
\begin{equation}
    \Lip_{\cS}(f)\ge C_p \frac{n}{d^{p-1}\deff(1)}.
\end{equation}
\end{proof}

\end{document}